\documentclass{article}
\usepackage{ijcai24}

\usepackage{times}
\usepackage{soul}
\usepackage{url}
\usepackage[hidelinks]{hyperref}
\usepackage[utf8]{inputenc}
\usepackage[small]{caption}
\usepackage{graphicx}
\usepackage{amsmath}
\usepackage{amsthm}
\usepackage{booktabs}
\usepackage[switch]{lineno}
\usepackage{dirtytalk}

\usepackage{hyperref}
\hypersetup{
    colorlinks=true,
    linkcolor=blue,
    filecolor=magenta,      
    urlcolor=cyan,
    citecolor=black,
    pdftitle={Interpretable Bayesian Optimization},
    pdfpagemode=FullScreen,
    }

\usepackage{mathtools} 
\usepackage{booktabs} 
\usepackage{tikz} 
\usepackage{amsmath}
\usepackage{amsthm}
\usepackage{amssymb}
\usepackage{bbm}
\usepackage{bm}
\usepackage[space]{cite}
\usepackage{float}
\usepackage[]{algorithm, algpseudocode} 

\DeclareMathOperator*{\argmin}{arg\,min}

\newcommand{\overbar}[1]{\mkern 1.5mu\overline{\mkern-1.5mu#1\mkern-1.5mu}\mkern 1.5mu}

\newtheorem*{axiom_efficiency}{Efficiency}
\newtheorem*{axiom_symmetry}{Symmetry}
\newtheorem*{axiom_dummy}{Dummy player}
\newtheorem*{axiom_linearity}{Linearity}

\newtheorem{Hypothesis}{Hypothesis}

\title{Explaining Bayesian Optimization by Shapley Values\\ Facilitates Human-AI Collaboration 

}

\author{
Julian Rodemann$^1$
\and
Federico Croppi$^{1}$\and
Philipp Arens$^2$\and
Yusuf Sale$^{3,4}$\and \\
Julia Herbinger$^{1,4}$\and 
Bernd Bischl$^{1,4}$\and
Eyke Hüllermeier$^{3,4}$\and
Thomas Augustin$^1$\and \\
Conor J. Walsh$^{2,5}$\And
Giuseppe Casalicchio$^{1,4}$\\
\affiliations
$^1$Department of Statistics, Ludwig-Maximilians-Universität (LMU) Munich\\
$^2$ John A. Paulson Harvard School of Engineering and Applied Sciences, Harvard University\\
$^3$Institute of Informatics, Ludwig-Maximilians-Universität (LMU) Munich\\
$^4$Munich Center for Machine Learning (MCML)\\
$^5$ Wyss Institute for Biologically Inspired Engineering, Harvard University\\
\vspace{0.1cm}
Correspondence to: \href{mailto:j.rodemann@lmu.de}{j.rodemann@lmu.de} 
}

\begin{document}

\maketitle

\begin{abstract}
Bayesian optimization (BO) with Gaussian processes (GP) has become an indispensable algorithm for black box optimization problems.
Not without a dash of irony, BO is often considered a black box itself, lacking ways to provide reasons as to why certain parameters are proposed to be evaluated. This is particularly relevant in human-in-the-loop applications of BO, such as in robotics. We address this issue by proposing \texttt{ShapleyBO}, a framework for interpreting BO's proposals by game-theoretic Shapley values. 
They quantify each parameter's contribution to BO's acquisition function. Exploiting the linearity of Shapley values, we are further able to identify how strongly each parameter drives BO's exploration and exploitation for additive acquisition functions like the confidence bound.
We also show that \texttt{ShapleyBO} can disentangle the contributions to exploration into those that explore aleatoric and epistemic uncertainty. Moreover, our method gives rise to a \texttt{ShapleyBO}-assisted human machine interface (HMI), allowing users to interfere with BO in case proposals do not align with human reasoning. We demonstrate this HMI's benefits for the use case of personalizing wearable robotic devices (assistive back exosuits) by human-in-the-loop BO. Results suggest human-BO teams with access to \texttt{ShapleyBO} can achieve lower regret than teams without.\footnote{Implementation of \texttt{ShapleyBO} and code to reproduce findings available at \url{https://anonymous.4open.science/r/ShapleyBO-E65D}.}
\end{abstract}

\section{Introduction}\label{sec:intro}

In artificial intelligence (AI) and machine learning (ML), the black-box nature of increasingly complex models poses serious challenges to end-users and researchers alike. As sophisticated algorithms have become more and more prevalent in critical aspects of society, the ability to comprehend their decision-making becomes paramount. The need for transparency and interpretability in AI systems is not merely an academic pursuit but a necessity for ensuring trust, accountability and ethical deployment of such technology.
The terms explainable AI (XAI) and interpretable machine learning (IML) are often used interchangeably
and describe efforts to help elucidate decision-making processes of intricate learning algorithms.
While the pursuit for interpretability is not a new endeavor, the widespread adoption of sophisticated ML models has intensified the need for IML methods. A broad spectrum of approaches has emerged in recent years, ranging from model-agnostic techniques to inherently interpretable model architectures, see \cite{molnar2020interpretable,molnar2020interpretablebook} for an overview.

This paper expands the focus of interpretability from ML models to black-box optimizers such as Bayesian optimization (BO) with Gaussian Processes (GPs). Optimization techniques of this kind find parameter configurations that optimize some unknown and expensive-to-evaluate target function by efficiently sampling from it. They play an important role in AI and are often perceived as black boxes themselves. Understanding and interpreting such optimization approaches can increase trust in systems that rely on them, mitigating algorithmic aversion \cite{dietvorst2015algorithm,burton2023beyond}. What is more, IML techniques can help speed up the optimization in a human-AI collaborative setup directly. Here, a human can intervene and reject or rectify proposals made by BO \cite{human-bo-team}. A better understanding of the algorithm fosters human-machine interaction which is key in such applications, as we demonstrate in this work. 

We focus on explaining Bayesian optimization. In particular, we will interpret proposed parameter configurations through Shapley values, a concept from cooperative game theory that has gained much popularity in IML. Our framework \texttt{ShapleyBO} informs users about how much each parameter contributed to the configurations proposed by a BO algorithm. The key idea is to quantify each parameters' contribution to the acquisition function instead of to the model's predictions, as would be customary in IML. Loosely speaking, the acquisition function describes how \say{informative} BO considers a given parameter configuration. A Shapley value can thus inform the user how much a single feature contributes to this informativeness. 
In each iteration, BO proposes parameters that maximize the acquisition function. By computing Shapley values for these points, we can inform BO users as to why this particular configuration has been selected over others.  
Since Shapley values are linear in the contributions they explain, see Axiom 3 in Section~\ref{background-shapley}, they can be used to inform us about how much each parameter contributed to each component of any additive acquisition function like the popular confidence bound. This turns out to be particularly helpful for disentangling parameters' contributions to exploration (uncertainty reduction) and exploitation (mean optimization) in Bayesian optimization. Trading off these two is fundamental to BO. 
While exploitation describes the incentive to greedily optimize based on existing secured knowledge, exploration is about risking to evaluate the target functions in regions where less prior knowledge is available in order to reduce uncertainty about those. 

In the theoretical part of this paper, we delve further into this distinction. In particular, we propose a method to further dissect the exploratory component into different kinds of uncertainties that a proposal seeks to reduce. 
Inspired by recent advancements in uncertainty quantification \cite{hullermeier2021aleatoric}, we suggest disentangling a parameter's exploratory contribution into the reduction of aleatoric and epistemic uncertainty. 
Aleatoric uncertainty (AU) emerges through inherent stochasticity of the data-generating process, omitted variables or measurement errors. 
Consequently, AU is a fixed, but unknown quantity. 
On the other hand, epistemic uncertainty (EU) is due to lack of knowledge about the best way to model the underlying data-generating process. As such, it is in principle reducible.  
Moreover, EU can be further subdivided into approximation uncertainty and model uncertainty, see \cite{hullermeier2021aleatoric}. 
While approximation uncertainty is well-described by BO's internal variance prediction, we deploy a set-valued prior near ignorance GP \cite{mangili2016prior,prior-robust-BO} to measure the model uncertainty. 

We further integrate \texttt{ShapleyBO} into human-AI collaborative BO. We propose a Human-Machine Interface (HMI) that allows users to better understand BO's proposals and utilize this understanding in deciding whether to intervene and rectify proposals or not. Experiments on data from a real-world use case of personalizing assistance parameters for a wearable back exosuit suggest that such an understanding can help to intervene more efficiently than without the availability of Shapley values, thus speeding up the optimization.
%

In summary, we make the following contributions. 

\textbf{(1)} We explain why parameters are proposed in Bayesian optimization by quantifying each parameters' contribution to a proposal through Shapley values.

\textbf{(2)} We further distinguish between parameters that drive exploitation (mean optimization) and exploration (uncertainty reduction) in Bayesian optimization, utilizing the linearity of Shapley values. 

\textbf{(3)} Exploratory uncertainty reduction is in turn dissected into aleatoric and epistemic sources of uncertainty, which fosters theoretical understanding of Bayesian optimization. 

\textbf{(4)} We apply our framework to simulated exosuit customization through human-in-the-loop Bayesian optimization and demonstrate that our method can speed up the procedure through more efficient human-machine interaction.


%


\section{Background}
\label{sec:background}
\subsection{Bayesian Optimization}
Bayesian optimization (BO) is a popular derivative-free optimizer for functions that are expensive to evaluate and lack an analytical description. Its origin dates back to \cite{mockus1975bayesian}. Modern use cases of BO cover engineering, drug discovery and finance as well as hyperparameter optimization and neural architecture search in ML, see e.g. \cite{frazier2016bayesian,pyzer2018bayesian,snoek2012practical} . BO approximates the target function through a surrogate model (SM). In the case of real-valued parameters, the latter typically is
a Gaussian process (GP). BO then combines the GP’s mean and standard error predictions to an acquisition function (AF), that is optimized in order to propose new points. 
Algorithm~\ref{algo-bo} summarizes Bayesian optimization applied to an optimization problem
\begin{equation}
    \min_{\bm{\theta} \in \bm \Theta} f(\bm{\theta}) \, .
\end{equation}
Observations of the target function $f$ are generated through
\begin{equation}\label{eq:targ-fun}
    \Psi: \bm \Theta \rightarrow \mathbb{R}, 
    \bm \theta \mapsto f(\bm \theta) + \bm \epsilon \, ,
\end{equation}
with $\bm \Theta$ a $p$-dimensional parameter space and $\bm \epsilon$ a zero-mean real-valued random variable. That is, we observe a noisy version $\Psi(\bm \theta)$ of continuous $f(\bm \theta)$. Here and henceforth, minimization is considered without loss of generality. We will further restrict ourselves to GPs as SMs and focus on the confidence bound (CB) as an AF, see Section~\ref{main}. In the human-in-the-loop setup, a user can intervene by either rejecting a proposal (line~\ref{algo-bo-af-std} in Algorithm~\ref{algo-bo}) or an update (line~\ref{algo-update} in Algorithm~\ref{algo-bo}) or by proposing another configuration, see Section~\ref{sec:application}.    
    
\begin{algorithm}[H]
\begin{center}
\caption{Bayesian Optimization}
\label{algo-bo}
    \begin{algorithmic}[1]
    \State create an initial design $D = \{(\bm{\theta}^{(i)}, \Psi^{(i)})\}_{i = 1,..., n_{init}}$ 
    \While {termination criterion is not fulfilled}
    \State \textbf{train} SM on data $D$\label{algo-bo-train-sm-std}
    \State \textbf{propose} $\bm{\theta}^{new}$ that optimizes $AF(SM(\bm{\theta}))$ \label{algo-bo-af-std}
    \State \textbf{evaluate} $\Psi$ on $\bm{\theta}^{new}$ 
    \State \textbf{update} $D \leftarrow D \cup {(\bm{\theta}^{new}, \Psi(\bm{\theta}^{new}))}$ \label{algo-update}
    \EndWhile
    \State \textbf{return} $\argmin_{\bm{\theta} \in D} \Psi(\bm{\theta})$
    \end{algorithmic}
    \end{center}
\end{algorithm}





\subsection{Shapley Values}
\label{background-shapley}

Shapley values are a concept from cooperative game theory, originally introduced by \cite{shapley1953value}, that can be used to measure the contribution of each feature to a prediction in an ML model. The key idea is to consider each feature as a player in a game where the prediction is the payoff, and to distribute the payoff fairly among the players according to their marginal contributions. Shapley values have several desirable properties 
that make them appealing for interpreting optimization problems. 
%
In general, given a set of players $P = \{1,\ldots, p\}$ and a value or payout function $v: 2^P \rightarrow \mathbb{R}$ that assigns a value $v(S)$ to every subset (called a \textit{coalition} in game theory) $S \subseteq P$ 
(such that $v(\emptyset) = 0$),
%
the Shapley value $\phi_{j}(v)$ of player $j$ is defined in terms of a weighted average of all its marginal contributions \cite{peters2015game}: 

\begin{equation}
\label{eq:sv_short}
    \phi_{j}(v) = \sum_{S\subseteq P \backslash\{j\}} 
                      \frac{|S|!\,(p-1-|S|)!}{p!}\,
                      [v(S\cup j) - v(S)] \;
\end{equation}
The Shapley value can be justified axiomatically through the properties of dummy player, efficiency, linearity, and symmetry, see e.g.\ \cite{croppi2021explaining}. 
\begin{axiom_dummy}
If $v(S \cup \{j\}) = v(S)$ for player $j$ and $\forall S \subseteq P \backslash \{j\}$,  then $\phi_j(v) = 0$.
\end{axiom_dummy}
\begin{axiom_efficiency}
$\sum_{j=1}^{p} \phi_{j}(v) = v(P) - v(\emptyset)$
\end{axiom_efficiency}
\begin{axiom_linearity}
Given two games $(P,v_1)$ and $(P,v_2)$ and any $a,b \in \mathbb{R}$, the following holds:
\[\phi_{j}(av_1 + bv_2) = a\phi_{j}(v_1) + b\phi_{j}(v_2)\]
\end{axiom_linearity}
\begin{axiom_symmetry}
If $v(S \cup \{j\})=v(S \cup \{l\})$ for players $j,l$ and every $S \subseteq P \backslash \{j,l\}$, then $\phi_j(v) = \phi_l(v)$.
\end{axiom_symmetry}

The contribution function does not require any specific properties and the Shapley value can hence be used in many different applications \cite[p.3]{frechette2015using}. Indeed it has recently become a state of the art method within the field of IML (see, e.g., \cite{lundberg2017unified, lundberg2018consistent}), but it has also been adapted in AutoML to explain algorithm selection problems \cite{frechette2015using}. Within the context of IML, Shapley values are considered a model agnostic local interpretation method. In particular, it breaks down an ML model's prediction into feature contributions for single instances. 
Compared to other IML methods such as permutation feature importance \cite{Breiman2001, fisher2019all} or partial dependence plot \cite{friedman2001greedy}, Shapley values have the main advantage that they fairly distribute feature interactions to quantify feature contributions. While the features become the players, 
the the contribution function is typically set to the expected output of the predictive model conditioned on the values of the features in a coalition, see \cite{lundberg2017unified, vstrumbelj2014explaining} or \cite[Equation 2]{aas2021explaining}. 
Formally, let $\hat{f}: \bm \Theta \rightarrow \mathbb{R}$ be a prediction model on feature space $\bm \Theta$ and $\tilde{\bm{\theta}} \in \bm \Theta$ the instance to explain. Then the worth of a coalition of features $S \subseteq \bm \Theta$ 
is given by $v(S) = \mathbb{E}[\hat{f}(\bm{\theta})| \bm \theta_S = \tilde{\bm{\theta}}_{S}]$, where $\bm \theta_S, \tilde{\bm{\theta}}_{S} \in S$ are the feature vectors $\bm \theta, \tilde{\bm{\theta}}$ projected onto $S$.

\subsection{Related Work}


As mentioned in Section~\ref{background-shapley}, there are only quite mild regularity conditions for a function to be explainable by Shapley values. Consequently, there exists a broad body of research on deploying Shapley values beyond classical prediction functions. Examples comprise the explanation of predictive uncertainty \cite{watson2023explaining}, reinforcement learning \cite{pmlr-v202-beechey23a}, or anomaly detection \cite{tallon2020explainable}. To the best of our knowledge, however, we are the first to explicitly explain Bayesian optimization through Shapley values, with the notable exception of concurrent work by \cite{adachi2024looping}, see below. Nevertheless, there is some work on Shapley-based explanations of other optimization algorithms such as evolutionary algorithms \cite{wang2023adapted}    
or differentiable architecture search (DARTS) in deep learning \cite{Xiao_2022_CVPR}. There are even efforts to utilize Shapley values to improve optimizers similar to our Shapley-assisted human-BO team. For instance, \cite{wu2023solving} solves fuzzy optimization problems by integrating Shapley values with evolutionary algortihms and \cite{bakshi2023shapley} use Shapley values to speed up multi-objective particle swarm optimization grey wolf optimization (PSOGWO).

Generally, there has been a lot of interest in how to incorporate human knowledge in optimization loops recently \cite{av2024enhanced,human-ai-collab,human-bo-team} and what role IML can play in this regard \cite{exai-human-ai-team}. This growing interest is not only sparked by fine-tuning large language models through reinforcement learning from human feedback \cite{rafailov2024direct}, but also by chemical applications \cite{cisse2023hypbo}. 
Very recently, \cite{adachi2024looping} introduced Collaborative and Explainable Bayesian Optimization (CoExBo) for lithium-ion battery design, a framework that integrates human knowledge into BO via preference learning and explains its proposals by Shapley values. Contrary to our approach, CoExBo first aligns human knowledge with BO by preference learning. In a second step, it then proposes several points and allows the user to select among them based on additionally provided Shapley values, while our Shapley-assisted human-BO team directly uses Shapley values to align a single BO proposal with human proposals.

\cite{chakraborty2023post,chakraborty2024explainable} recently proposed TNTRules, a post-hoc rule-based explanation method of Bayesian optimization. TNTRules finds (through clustering algorithms) subspaces of the parameter space that should be tuned by the user. Similar to our work, it emphasizes the benefits of XAI methods in human-collaborative BO. Contrary to our work, it is a post-hoc method (\texttt{ShapleyBO} works online) and focuses on explaining the parameter space rather than single BO proposals.

\section{Interpreting Bayesian Optimization through Shapley Values}
\label{main}

In this section we introduce \texttt{ShapleyBO}, a modular framework partly based on \cite{croppi2021explaining}, that allows to interpret BO proposals by Shapley values: 
Transitioning from ML models to acquisition functions, the utilization of the Shapley value becomes remarkably straightforward, as an acquisition function essentially represents a transformed version of a surrogate (ML) model's prediction function. 
Consequently, the Shapley value can be employed with any acquisition function to evaluate the informativeness of selected parameter values.
Among an array of AF options, the confidence bound \cite{auer2002finite,kocsis2006bandit} appears particularly suited for our approach, due to its intuitive functional form and additive nature. 
The \textbf{confidence bound (CB)} of a parameter vector $\bm{\theta} \in \bm \Theta$ is defined as
\begin{equation}
\label{def:lcb}
    cb(\bm{\theta}) = \hat{\mu}(\bm{\theta}) -\lambda\hat{\sigma}(\bm{\theta}), 
\end{equation}
where $\hat \mu$ and $\hat \sigma$ are mean and standard error estimates by the surrogate model (here: GP), respectively; $\lambda > 0$ is a hyperparameter controlling the exploration-exploitation trade-off.
For ease of exposition, we will henceforth refer to the confidence bound as $m -\lambda se$. The rationale behind the confidence bound is fairly intuitive: a point is deemed desirable if either (i) the mean prediction $m$ is low (indicating an anti\-cipation of a low target value, thus \textit{exploiting} existing knowledge) or (ii) the uncertainty prediction $se$ is high, indicating limited information about the target function in that area (\textit{exploring}). 
\begin{figure}[t]
    \centering
    \includegraphics[scale = 0.26]{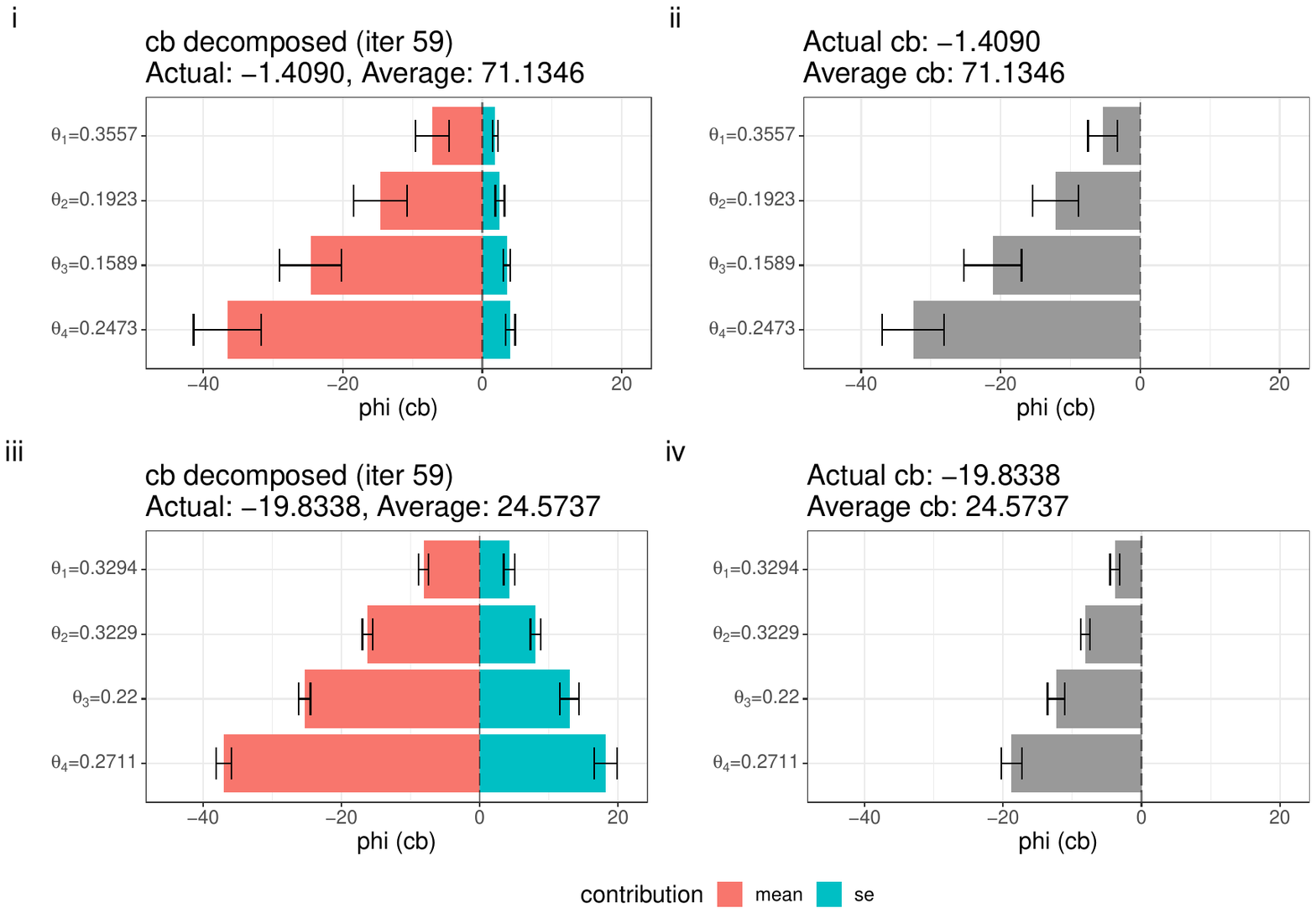}
    \caption[Hyper-Ellipsoid: contributions in iteration 59 of noise-free example]{\texttt{ShapleyBO} results in iteration 59 of BO on $\Psi(\bm \theta) = f(\bm \theta) + \bm \epsilon$. Plots i and ii for $\lambda = 1$ and plots iii and iv for $\lambda = 10$, see \cite[Fig. 5]{croppi2021explaining}. Contributions ($phi$) are averaged over 30 proposals for each $\lambda$. On the right, the overall informativeness of the parameters is displayed ($cb$ contributions), and on the left the decomposition into $m$ (red) and $se$ (blue) contributions. Recall that cb is minimized. Error bars: Standard deviation of estimates. 
    }
    \label{fig:he_iter59}
\end{figure}
\begin{figure}[t]
    \centering
    \includegraphics[scale = 0.26]{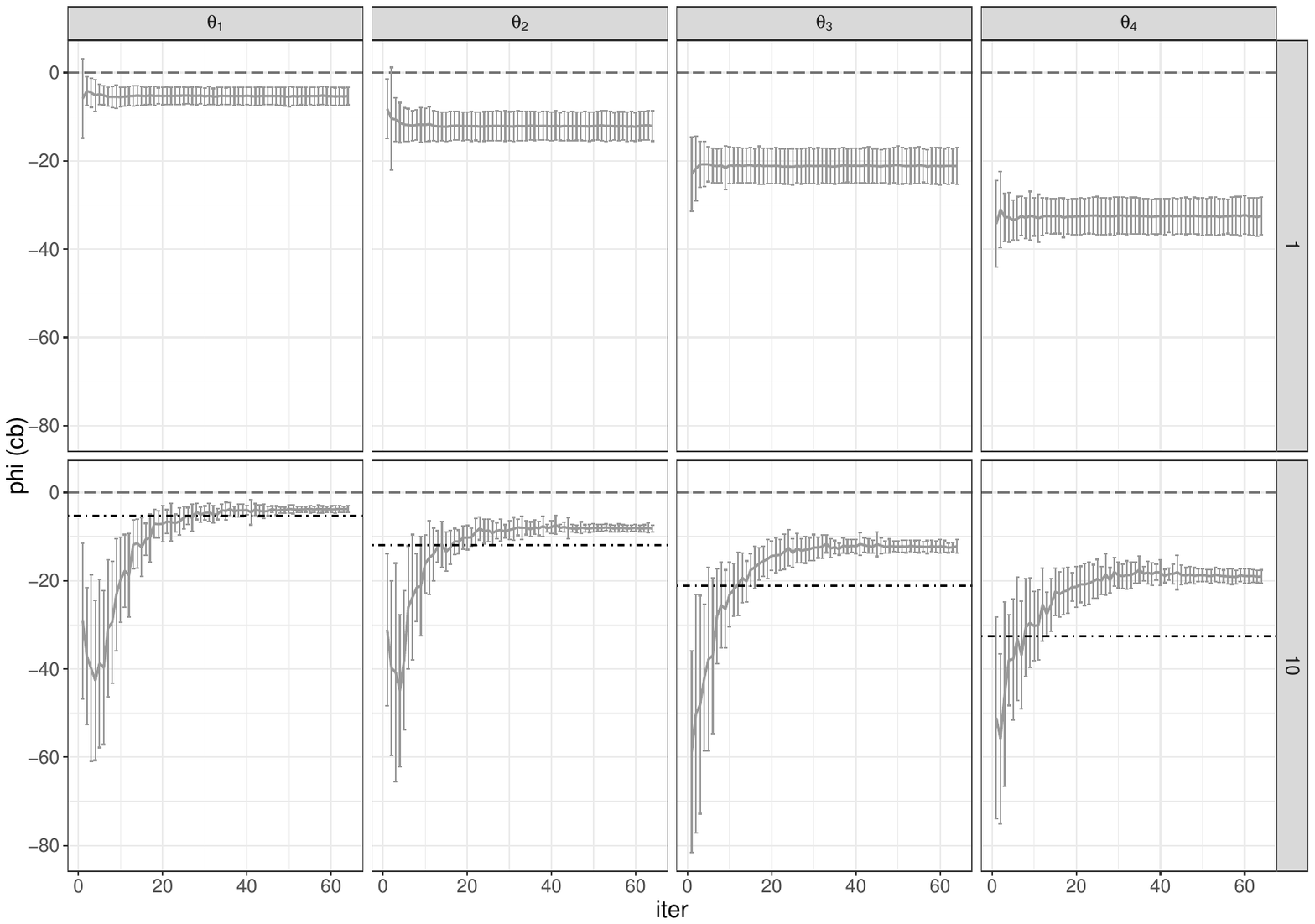}
    \centering
    \includegraphics[scale = 0.26]{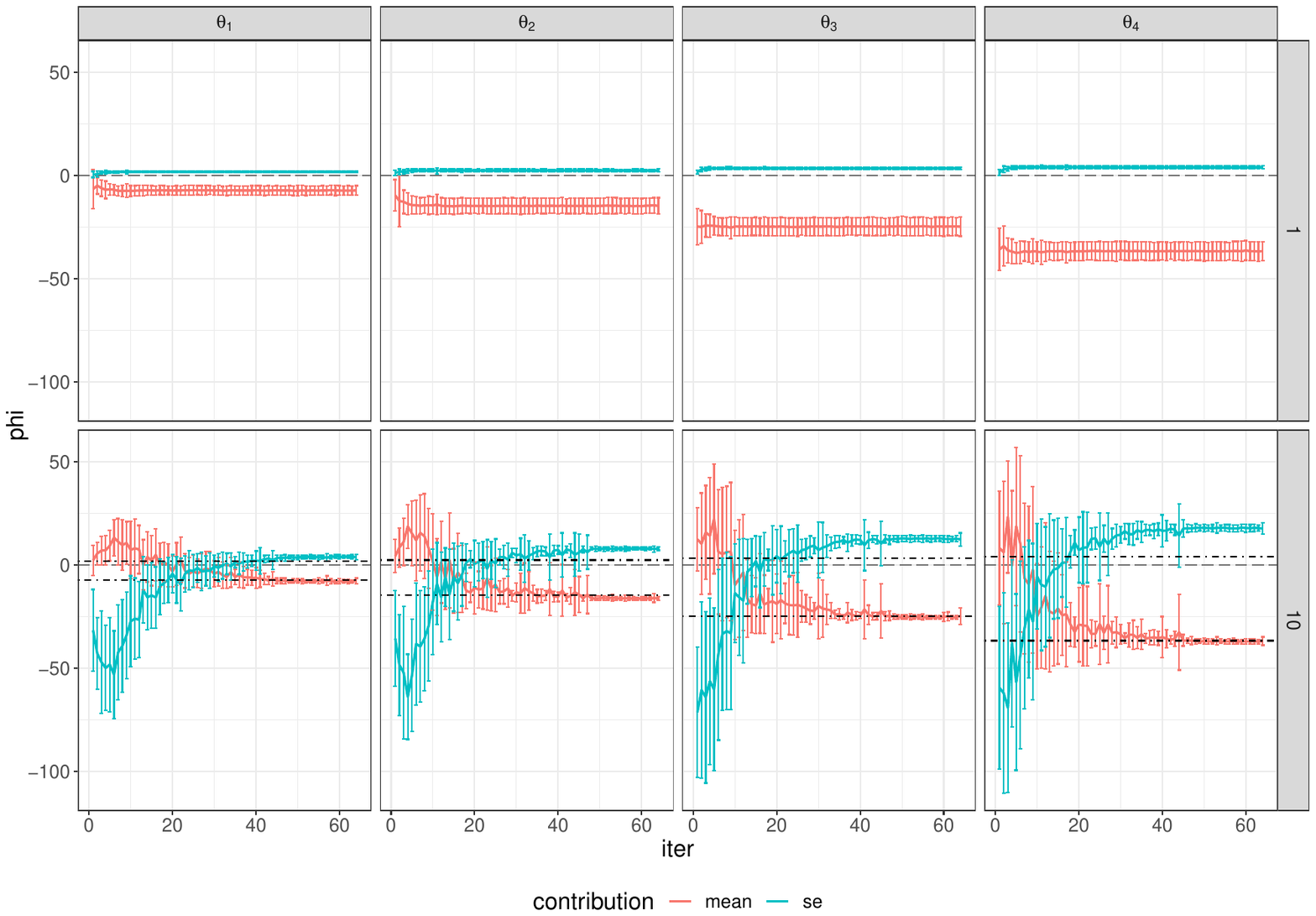}
    \caption[Hyper-Ellipsoid: desirability paths]{Informativeness paths for hyper ellipsoid optimization, see \cite[Fig. 6]{croppi2021explaining}. Plot on top displays $cb$ contributions with facets for parameters (vertical) and $\lambda$ (horizontal); beneath its decomposition into $m$ (red) and $se$ (blue) contributions. They are averaged over 30 proposals in each iteration and the uncertainty of the estimate is displayed with error bars using one standard deviation. The black dot-dashed line in the $\lambda = 10$ plots displays the average contribution of the parameters in the $\lambda = 1$ run.} 
    \label{fig:he_des_paths}
\end{figure}
Proposing new samples boils down to optimizing this confidence bound. To this end, let minimizing (w.l.o.g.) the confidence bound be a cooperative game along the lines of Section~\ref{background-shapley}. It shall be defined as $(P, cb)$, or as two separate games $(P, m)$ and $(P, se)$, with $P$ being the grand coalition of Shapley values involved and  $m$ and $se$ the mean and uncertainty prediction of the Shapley values, respectively. According to the linearity axiom the $cb$ contribution of any parameter $j$ of explicand $\tilde{\bm{\theta}}$ can then be decomposed into mean contribution $\phi_{j}(m)$ and uncertainty contribution $\phi_{j}(se)$.

\begin{equation}
\label{eq:sv_lcb}
    \phi_{j}(cb) = \phi_{j}(m - \lambda se) = \phi_{j}(m) - \lambda\phi_{j}(se)
\end{equation}

Thus, we can not only evaluate the overall informativeness of each parameter $\theta_j$, but also examine how both contributions $\phi(m)$ and $\phi(se)$ impact and drive the selection of proposed parameter values, shedding some light on the exploration-exploitation trade-off. 
On the background of recent work on uncertainty quantification \cite{mckeand2021stochastic,hullermeier2021aleatoric,psaros2023uncertainty,pmlr-v216-jansen23a}, we go beyond \cite{croppi2021explaining} and further aim at disentangling the uncertainty contribution $\phi_{j}(se)$ of a parameter $\theta_j$ into its epistemic (reducible) and aleatoric (irreducible) part. Aleaotoric uncertainty is typically caused by noise. The latter is particularly relevant in Bayesian optimization if it is of heteroscedastic nature, i.e., dependent on $\bm \theta$, since decision makers are often risk-averse. In other words, when deciding among two parameter configurations with equal mean target, most humans tend to opt for the one with lower variation. This motivates a risk-averse optimization problem: $\min_{\bm{\theta} \in \bm \Theta} f(\bm \theta) - \alpha \cdot \epsilon(\bm \theta)$ with $\epsilon(\bm \theta)$ some noise that is non-constant in $\bm \theta$.     
\cite{makarova2021risk} propose risk-averse heteroscedastic Bayesian optimization (RAHBO) which entails minimizing (w.l.o.g.) the \textbf{risk-averse confidence bound (racb)} 

\begin{equation}\label{eq:racb}
  racb(\bm \theta) = \widehat\mu(\bm \theta) -  \tau  \cdot \widehat \sigma(\bm \theta) + \alpha  \cdot \widehat \epsilon(\bm \theta),
\end{equation}
where $\widehat \epsilon(\bm \theta)$ is an on-the-fly estimate of the noise. Due to $racb$'s additive structure, \texttt{ShapleyBO} can identify each parameter's contribution to epistemic uncertainty \textit{reduction} through $-  \tau  \cdot \widehat \sigma(\bm \theta)$ and to aleatoric uncertainty \textit{avoidance} through $\alpha  \cdot \widehat \epsilon(\bm \theta)$. By filtering out these exploratory contributions, the remainder of a parameter's overall Shapley value can be identified as the parameter's contribution to mean optimization through $\widehat\mu(\bm \theta)$ (exploitation). 
We extend RAHBO by further dissecting the epistemic uncertainty into model uncertainty and approximation uncertainty. While the former describes uncertainty arising from model choise, the latter refers to classical statistical sampling uncertainty \cite[section 2.4]{hullermeier2021aleatoric}. Notably, the model uncertainty asymptotically vanishes for GPs with universal approximation property, as the likelihood dominates the posterior in the limit, see \cite{mangili2016prior}. For the finite sample regime in BO, however, this asymptotic behavior is not relevant. In order to quantify model uncertainty, we rely on recently introduced prior-mean robust Bayesian optimization (PROBO) \cite{rodemann2021robust,rodemann2021accounting-isipta,prior-robust-BO}.
PROBO avoids GP prior mean parameter misspecification by deploying imprecise Gaussian processes \cite{mangili2016prior} as surrogate models. The rough idea of PROBO is to specify a set of prior mean values resulting in a set of posterior GPs. These latter give rise to an upper and lower mean estimate $\overline{\hat \mu}(\bm \theta)_c$ and $\underline{\hat \mu}(\bm \theta)_c$ subject to a degree of imprecision $c$.
We define the \textbf{uncertainty aware confidence bound (uacb)} as follows: $uacb(\bm \theta,\lambda,\rho,\alpha) =$ 
\vspace{0.2cm}


\begin{equation}
\resizebox{.9\hsize}{!}{$
          \widehat\mu(\bm \theta) \; -  \underbrace{\underbrace{\lambda  \cdot \widehat{\sigma}(\bm \theta)}_\text{{approximation uncertainty}} -  \underbrace{\rho \cdot (\overline{\hat \mu}(\bm \theta)_c - \underline{\hat \mu}(\bm \theta)_c)}_\text{model uncertainty}}_\text{{epistemic uncertainty}} +  \underbrace{\alpha  \cdot \widehat{\mathbf{\epsilon}}(\bm \theta).}_\text{aleatoric uncertainty}  
          $}
\end{equation}

\vspace{0.2cm}

Since the remaining epistemic uncertainty is represented by the model's $\widehat{\sigma}(\bm \theta)$, we can interpret it as (an upper bound of) approximation uncertainty. Note that the uncertainty measured by PROBO is only a lower bound of the total model uncertainty, which also includes, for instance, uncertainty w.r.t. kernel parameter selection. The $uacb$ allows us to analyze as to what degree a parameter configuration is motivated by exploitation on the one hand or by reducing approximation, model or aleatoric uncertainty on the other hand. Since imprecise Gaussian processes are currently limited to univariate functions \cite{mangili2016prior}, we are not able to further attribute the contributions to multiple parameters. Future work will aim at multivariate extension. 
Nevertheless, the $uacb$ facilitates interpretation of proposals' contributions to all these uncertainty types by means of \texttt{ShapleyBO}.




\section{Illustrative Application}
\label{sec:illustration}

A deployment on synthetic functions allows us to validate our method, because we can formulate concrete expectations for the contributions based on the known functional form of the synthetic target function. 
We select a hyper-ellipsoid function, see Equation \ref{eq:ellipsoid-no-noise} below, where the parameters' partial derivatives grow in $j$. Thus, we expect \texttt{ShapleyBO} to identify parameters with high $j$ to be most influential in BO.

\textbf{Homoscedastic Target Function:}
Firstly, we illustrate \texttt{ShapleyBO} by optimizing $\Psi(\bm \theta) = f(\bm \theta) + \bm \epsilon$, where 

\begin{equation}
\label{eq:ellipsoid-no-noise}
 f: [-5.12, 5.12]^4 \rightarrow \mathbb R_0^{+}; \bm \theta \mapsto  f(\mathbf{\bm \theta}) = \sum_{j=1}^{4} j\cdot\mathbf{\theta}_j^{2} 
\end{equation}
with $j \in \{1,2,3,4\}$ and one unique optimum at $\bm{\theta}^{\ast} = (0,0,0,0)^T$ with $f(\bm{\theta}^{\ast}) = 0$. The noise $\epsilon$ is assumed be \textit{i.i.d.} Gaussian and independent of $\bm{\theta}$ (homoscedastic).
To control for the stochastic behavior of BO, $30$ optimizations with $60$ iterations each are run. After each optimization, \texttt{ShapleyBO} is applied. Results in each iteration are then averaged over all runs. 
Figure~\ref{fig:he_iter59} has the results exemplarily for iteration 59. They are displayed for $\lambda=1$ (i and ii) and $\lambda = 10$ (iii and iv). The analysis is based on \cite[section 7.1]{croppi2021explaining}

As expected in light of the partial derivatives of $f(\bm \theta)$, the contribution of $\theta_j$ grows with $j$. 
In contrast, uncertainty exhibits a diminutive and adverse effect. The uncertainty measurement $\hat{se}$ for the recommended setup falls below the mean $\bar{se}$, thus yielding a positive payout (negative contributions). Opting for a setup with an uncertainty estimate beneath the average is deemed a strategic compromise towards enhancing mean values at the expense of exploration. \texttt{ShapleyBO} facilitates a nuanced allocation of this trade-off across parameters, see both Figures~\ref{fig:he_iter59} and~\ref{fig:he_des_paths}.
We will also study how the contributions change in the course of the optimization. Respective informativeness paths are shown in Figure~\ref{fig:he_des_paths}.
They reveal consistent patterns for $\lambda = 1$, with a preference for higher parameters persisting across the board. 
Upon examining the lower plots' decomposition, we see that both mean and uncertainty contributions remain constant over time. With a setting of $\lambda = 10$, the algorithm conducts a thorough exploration of the parameter space, resulting in initially higher contributions and a more uniform informativeness among parameters. Over time, these contributions diminish smoothly, showing differences between parameters akin to the $\lambda = 1$ scenario, but with a notable decrease in informativeness for higher parameters.
Throughout the optimization process, the emphasis shifts from reducing uncertainty to prioritizing mean reduction, leading BO to favor configurations that perform well over those with high uncertainty. This transition is marked by a crossing in the contribution curves, see Figure~\ref{fig:he_des_paths}, indicating a preference for mean reduction over uncertainty reduction, particularly for parameters with higher $\theta_j$, underscoring their importance for the optimization.


\begin{figure}[t]
\begin{minipage}{0.49\linewidth}
        \includegraphics[scale=0.26, trim={0 0 0 0.8cm},clip]{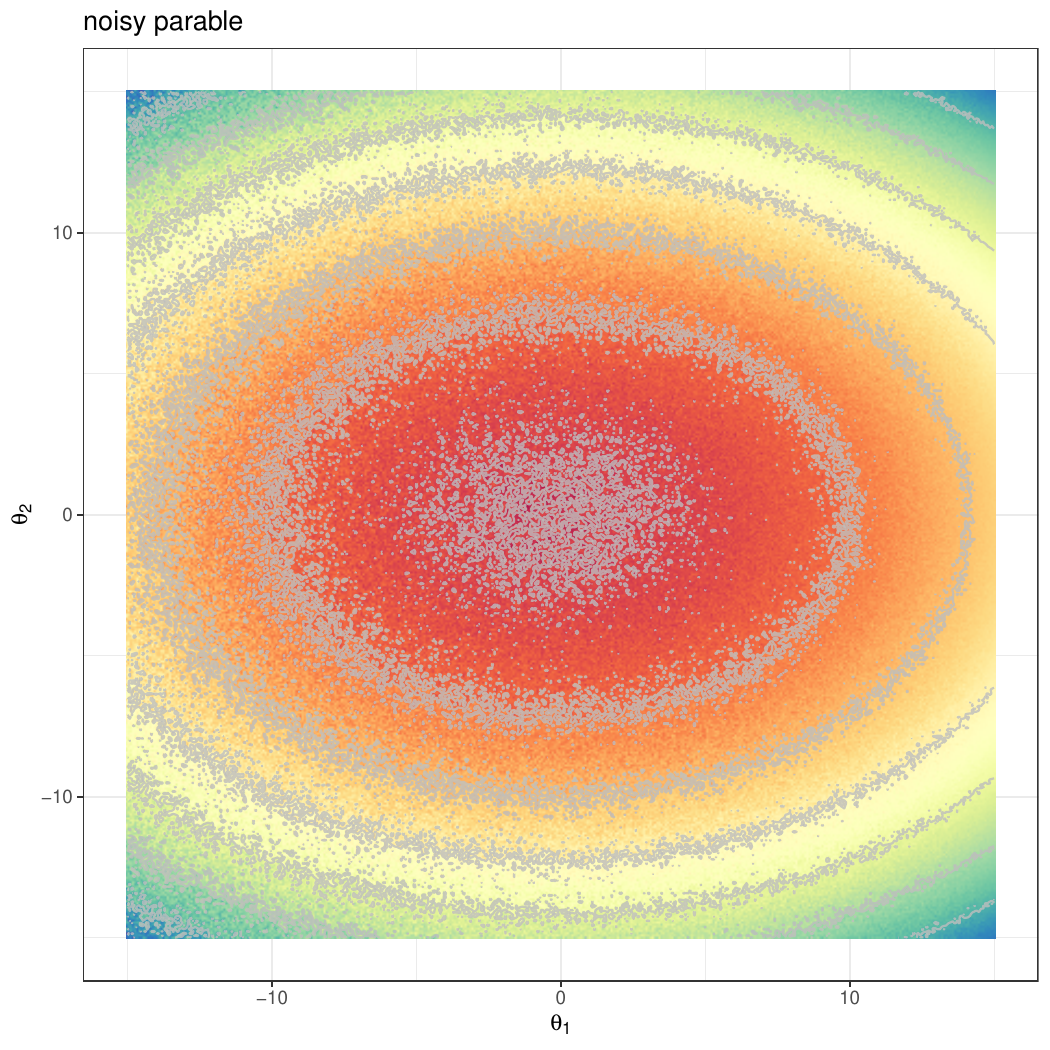}
\end{minipage}
\begin{minipage}{0.49\linewidth}
        \includegraphics[scale=0.26, trim={0 0 0 0.8cm},clip]{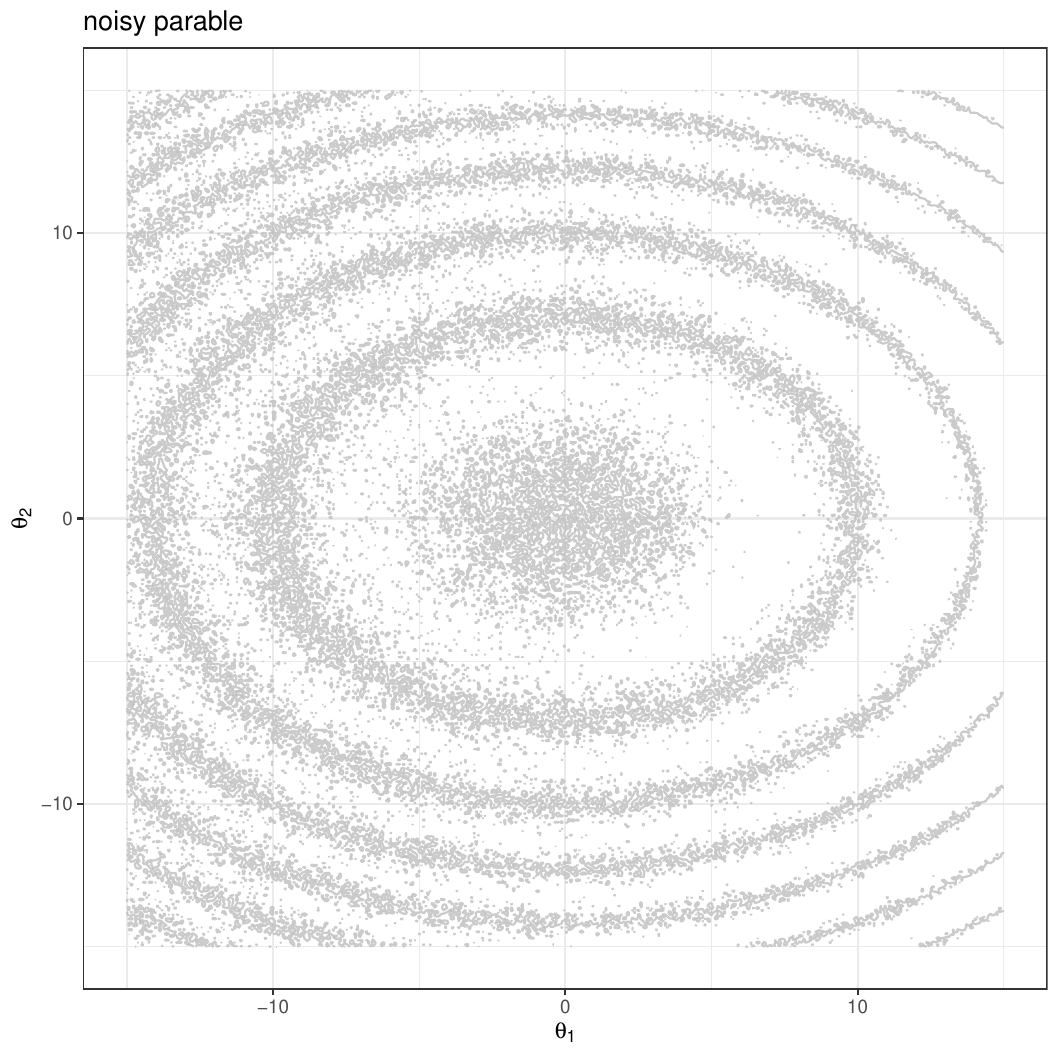}
\end{minipage}

    \caption{Contour plots of noisy ellipsoid function $\bm \Psi(\bm \theta) = g(\bm \theta) + \bm \epsilon(\bm \theta)$, see Equations~\ref{eq:ellipsoid}~and~\ref{eq:ellipsoid-noise}. Red: low $\bm \Psi(\bm \theta) $; blue: high $\bm \Psi(\bm \theta) $.}
        \label{fig:ellipsoid}
\end{figure}    

\textbf{Heteroscedastic Target Function:}
Secondly, we illustrate \texttt{ShapleyBO} for Bayesian optimization of a two-dimensional ellipsoid function with noise depending on $\bm \theta$. That is, we minimize $\bm \Psi(\bm \theta) = g(\bm \theta) + \bm \epsilon(\bm \theta)$, where
    \begin{equation}
    \label{eq:ellipsoid}
    g \colon [-15, 15]^2 \to \mathbb{R}_{0}^{+}; \;
    \bm{\theta} \mapsto f(\bm \theta) = \sum_{i}^{2} i \cdot \theta_i^2  
    \end{equation}    
with $i \in \{1, 2\}$ and     
    \begin{equation}
    \label{eq:ellipsoid-noise}
     \bm{\epsilon}(\bm \theta) = 30 \cdot \lvert \theta_1 - 15 \rvert + 0.3 \cdot \lvert \theta_2 - 15 \rvert.  
    \end{equation}

That is, the noise grows strongly in $\theta_1$, but only moderately in $\theta_2$. Figure~\ref{fig:ellipsoid} shows contours of $g(\bm \theta)$. It becomes evident that the function varies stronger w.r.t. $\theta_2$ than w.r.t. $\theta_1$, while the noise is strongly affected by $\theta_1$ and almost constant in $\theta_2$.  
Hence, we expect the respective Shapley values for aleatoric uncertainty contributions to be high for $\theta_1$ and low for $\theta_2$, and vice versa for exploitation (mean optimization). 
We run BO on $g(\bm \theta)$ with risk-averse confidence bound ($racb$), see Equation~\ref{eq:racb}; we again average over 30 restarts of BO with 60 iterations each. \texttt{ShapleyBO} delivers contributions for each $\theta_j$ to each of $racb$'s components in each of BO's iterations. Table~\ref{tab:shapleys-examp} has the results for exemplary iterations 1 and 2.
Table~\ref{tab:shapleys-hetero-avg} shows the contributions averaged over all $i \in \{1, \dots, 60\}$ iterations and all $r \in \{1,\dots, 30\}$ restarts. For instance, the averaged mean contributions of parameter $j$ are $\overbar{\phi_j}(m) = \frac{1}{30} \sum_{r=1}^{30} \frac{1}{60} \sum_{i=1}^{60} \phi_{j,r,i}(m)$. It becomes evident that $\theta_2$ is more important for the mean minimization than $\theta_1$, while the latter contributes more to aleatoric uncertainty (noise) avoidance. The contributions to epistemic uncertainty reduction vary less with the parameters.

Summing up, the applications on both homo- and heteroscedastic target functions demonstrated that \texttt{ShapleyBO} manages to disentangle contributions of different parameters to different objectives of BO, thus providing valuable insights both into BO's inner working (see Figures~\ref{fig:he_iter59},~\ref{fig:he_des_paths} and Table~\ref{tab:shapleys-examp}) and about the target function itself (see Table~\ref{tab:shapleys-hetero-avg}).

\begin{table}[]
    \centering
    \scalebox{0.95}{
\begin{tabular}{cccccc}
\hline iteration & $j$ & $\phi_j(m)$ & $\phi_j(se)$ & $\phi_j(\hat \epsilon)$ & $\phi_j(racb)$ \\
\hline 1 & $\theta_1$ & -100.2 & 2.4 & -13.9 & -111.8 \\
 1 & $\theta_2$ & -163.1 & 2.2 & 1.5 & -159.4 \\
 2 & $\theta_1$ & -87.8 & 2.4 & -37.7 & -123.1 \\
 2 & $\theta_2$ & -165.6 & 1.6 & 3.3 & -160.3 \\
\hline
\end{tabular}
}
    \caption{Results of \texttt{ShapleyBO} for exemplary iterations $1$ and $2$ of BO with risk-averse confidence bound (Equation~\ref{eq:racb}) on heteroscedastic target function $g(\bm \theta)$ (Equation~\ref{eq:ellipsoid-noise}).  }
    \label{tab:shapleys-examp}
\end{table}

\begin{table}[t]
    \centering
    \scalebox{0.95}{
    \begin{tabular}{cccc}
$j$ & $\overbar{\phi_j}(m)$ & $\overbar{\phi_j}(se)$ & $ \overbar{\phi_j}(\hat \epsilon)$  \\
    \hline
1 & -48.48 & 4.20 & -87.27\\
2 & -157.73 & 6.88 & 0.24 \\
\hline
    \end{tabular}
    }
    \caption{Results of \texttt{ShapleyBO} for averaged over all 60 iterations and all 30 BO restarts with risk-averse confidence bound (Equation~\ref{eq:racb}) on heteroscedastic target function $g(\bm \theta)$ (Equation~\ref{eq:ellipsoid-noise}).}    \label{tab:shapleys-hetero-avg}
\end{table}

\section{Shapley-Assisted Human Machine Interface}
\label{sec:application}
The ability to interpret Bayesian optimization can be particularly useful for Human-In-the-Loop (HIL) applications, where users observe each step in the sequential optimization procedure. In this case \texttt{ShapleyBO} can inform users online, that is while the optimization is still running, about why certain actions were taken over others, instead of providing such explanations after the experiment has concluded. More specifically, we consider a human-AI collaborative framework \cite{chakraborty2024explainable, gupta2023bo, human-bo-team, human-ai-collab, borji2013bayesian}, in which users can actively participate in the optimization by rejecting BO proposals and instead take actions on their own. As demonstrated in Section~\ref{sec:illustration}, Shapley values can provide structural insights on the relative importance of parameters for the optimization by filtering out uncertainty contributions, see $\overbar{\phi_j}(m)$ in Table~\ref{tab:shapleys-hetero-avg} for instance. Our general hypothesis is that basing the decision to intervene on this information will speed up the optimization. The underlying idea is that users can reject proposals in case the respective Shapley values do not align with the user's knowledge about the optimization problem.
To test this hypothesis, we benchmark a \texttt{ShapleyBO}-assisted human-AI team against teams without access to Shapley values. To better illustrate this, we consider the real-world use case of personalizing control parameters of a wearable, assistive back exosuit by Bayesian optimization.

\subsection{Personalizing Soft Exosuits}
\label{sec:exosuits}


\begin{figure}[t]
    \centering
    \includegraphics[width=0.99\linewidth]{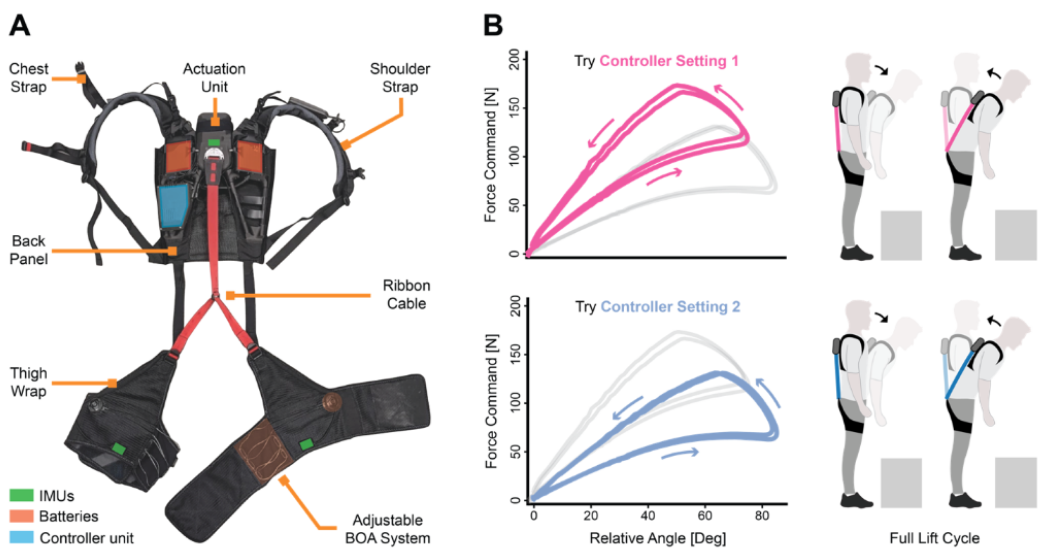}
    \caption{A: Assistive soft back exosuit. B: Force profile example for preference learning. Subjects are asked to compare controllers setting 1 (pink) to 2 (blue). Each option varies in the amount of lowering gain ($\theta_{low}$) and lifting gain ($\theta_{lif}$), see \cite{Arens2023Preference}.}
    \label{fig:exosuit}
\end{figure}

Wearable robotic devices, such as exoskeletons and exosuits, have emerged as promising tools in mitigating risk of injury and aiding rehabilitation \cite{siviy2023opportunities, toxiri2019back}. With an increase in use cases and accessibility to a broader community, it has become apparent that the benefits of such devices can vary substantially between individuals. Besides design choices, which have to be made early on and are therefore often guided by (average) user anthropometrics, important factors influencing device efficacy are the magnitude and timing of assistance. 
To understand which settings work best for an individual, many studies follow HIL frameworks. These approaches comprise a feedback loop in which the impact of a controller modification on the objective function of interest 
is measured in real-time, and used to determine a set of control parameters that are likely improve upon the current optimum in the subsequent iteration. 
Given that under such conditions there is typically no known analytical relationship between control inputs and objective function outputs, sample efficient, query based methods like Bayesian Optimization have had considerable success for such applications \cite{zhang2017human, ding2018human}. 

\subsection{Experimental Setup}


Here, we explore the potential of \texttt{ShapleyBO} for the use-case of preference-based assistance optimization for a soft back exosuit, see Figure~\ref{fig:exosuit}. To this end we consider a dataset in which 15 healthy individuals performed a simple, stoop lifting task with a light (2kg) external load \cite{Arens2023Preference}. 
Preference was queried in a forced-choice paradigm. That is, within each iteration, participants were consecutively exposed to two control parameter settings and asked to indicate which of the two options they preferred for completing the given task. Each of the settings comprised two parameters, referred to as lowering gain $\theta_{low}$ and lifting gain $\theta_{lif}$, which govern the amount of lowering and lifting assistance provided by the device, respectively, see also Figure~\ref{fig:exosuit}. 
This preference feedback was used to compute a posterior utility distribution over the considered parameter domains, relying on a probit likelihood model and a GP prior over the latent user utility as described in \cite{chu2005preference}. The experiment comprised three separate optimization blocks, in each of which the optimization was running for 12 iterations. To test \texttt{ShapleyBO} on this dataset, we averaged the three utility functions for each participant and interpolated by another GP to simulate the user's ground truth utility function.
The remaining setup in our experimental study closely follows the one in \cite{human-bo-team}. 
That is, the human can intervene in Bayesian optimization by rectifying proposals made by the algorithm, see pseudo code in Algorithm~\ref{algo-bo_human}.     
We will compare our \texttt{ShapleyBO}-assisted human-BO team against the team in \cite[Algorithm 1]{human-bo-team} and three other baselines (human alone, BO alone, human-BO team with different intervention criterion). We model human decisions by another Bayesian optimization, following \cite{borji2013bayesian,human-bo-team}. This means that $\bm \theta^{human} = (\theta^{human}_{lif},\theta^{human}_{low})^T$ is found by optimizing an acquisition function modeling human preferences. We use a BO with the same surrogate model and acquisition function as for the outer loop, but with different exploration-exploitation preference and different initial design, representing differing risk-aversion and knowledge of the human, respectively.
\setcounter{algorithm}{0}
\begin{algorithm}

\begin{center}
\begin{small}    
\caption{Human-AI Collaborative BO}
\label{algo-bo_human}

    \begin{algorithmic}[1]
    \State create an initial design $D = \{(\bm{\theta}^{(i)}, \Psi^{(i)})\}_{i = 1,..., n_{init}}$ 
    \While {termination criterion is not fulfilled}
    \State \textbf{train} SM on data $D$
    \State \textbf{propose} $\bm{\theta}^{new}$ that optimizes $AF(SM(\bm{\theta}))$ 
    \If {intervention criterion is fulfilled} \label{alg:intervent-crit}
    \State $\bm{\theta}^{new} \leftarrow \bm{\theta}^{human}$ 
    \EndIf
    \State \textbf{evaluate} $\Psi$ on $\bm{\theta}^{new}$ 
    \State \textbf{update} $D \leftarrow D \cup {(\bm{\theta}^{new}, \Psi(\bm{\theta}^{new}))}$ 
    \EndWhile
    \State \textbf{return} $\argmin_{\bm{\theta} \in D} \Psi(\bm{\theta})$ 
    \end{algorithmic}
\end{small}
    \end{center}
\end{algorithm}
\begin{table}[]
  \scalebox{.6}{
    \centering
    \begin{tabular}{c|ccccc}
        Agent & A0 & A1 & A2 & A3 & A4  \\
         & BO & Human & Param-Team & \cite{human-bo-team} & Shap-Team \\
         \hline
        Intervention &&&&& \\
        Criterion & never  & always & $\theta^{new}_{lif},\theta^{new}_{low}$ & $k$-th iteration & $\phi_{lif}^{new}(m), \phi_{low}^{new}(m)$

    \end{tabular}
    }
    \caption{All agents: \texttt{ShapleyBO}-assisted A4 and baselines A0-A3. }
    \label{tab:my_label}
\end{table}
All agents (A0, A1, A2, A3, A4) are equal to each other, the only difference being that the \texttt{ShapleyBO}-assisted agent intervenes based on Shapley values (A4), while the other agents intervene in each $k$-th iteration (A3) \cite{human-bo-team}, based on the proposed parameters (A2), always (A1) or completely abstain from intervening (A0), see overview in Table~\ref{tab:my_label}. 
A4 has access to \texttt{ShapleyBO} and bases their decision to intervene (line~\ref{alg:intervent-crit} in Algorithm~\ref{algo-bo_human}) on the alignment of the Shapley values of a BO proposal $ \bm \theta^{new} = (\theta^{new}_{lif}, \theta^{new}_{low})$ with the agent's knowledge. 
More precisely, A4 accepts a BO proposal $\theta$ (does not intervene) if 
\begin{equation}
    \frac{1}{\beta} < \frac{\phi_{lif}^{new}(m)}{\phi_{low}^{new}(m)} / \frac{1}{T} \sum_{t=1}^T \frac{\phi_{lif}^{human}(m)_t}{\phi_{low}^{human}(m)_t} < \beta,   
\end{equation}
where $t \in \{1, \dots, T\}$ are iterations of the BO modeling the agent and $\phi(m)$ the Shapley mean contributions of $(\theta_{lif}, \theta_{low})$, i.e., the exosuit's lifting and lowering gain, respectively. We discuss different Shapley-based intervention criteria in the supplement. For A2's intervention criterion we consider the alignment of $(\theta^{new}_{lif}, \theta^{new}_{low})$ with the agents knowledge based on the parameter values itself. That is, A3 accepts a BO proposal (does not intervene) if
\begin{equation}
    \frac{1}{\beta} < \frac{\theta^{new}_{lif}}{\theta^{new}_{low}} / \frac{1}{T} \sum_{t=1}^T \frac{\theta_{lif,t}^{human}}{\theta_{low,t}^{human}} < \beta. 
\end{equation}



\subsection{Results}

\begin{figure}[]
    \centering
    \includegraphics[width=.999\linewidth]{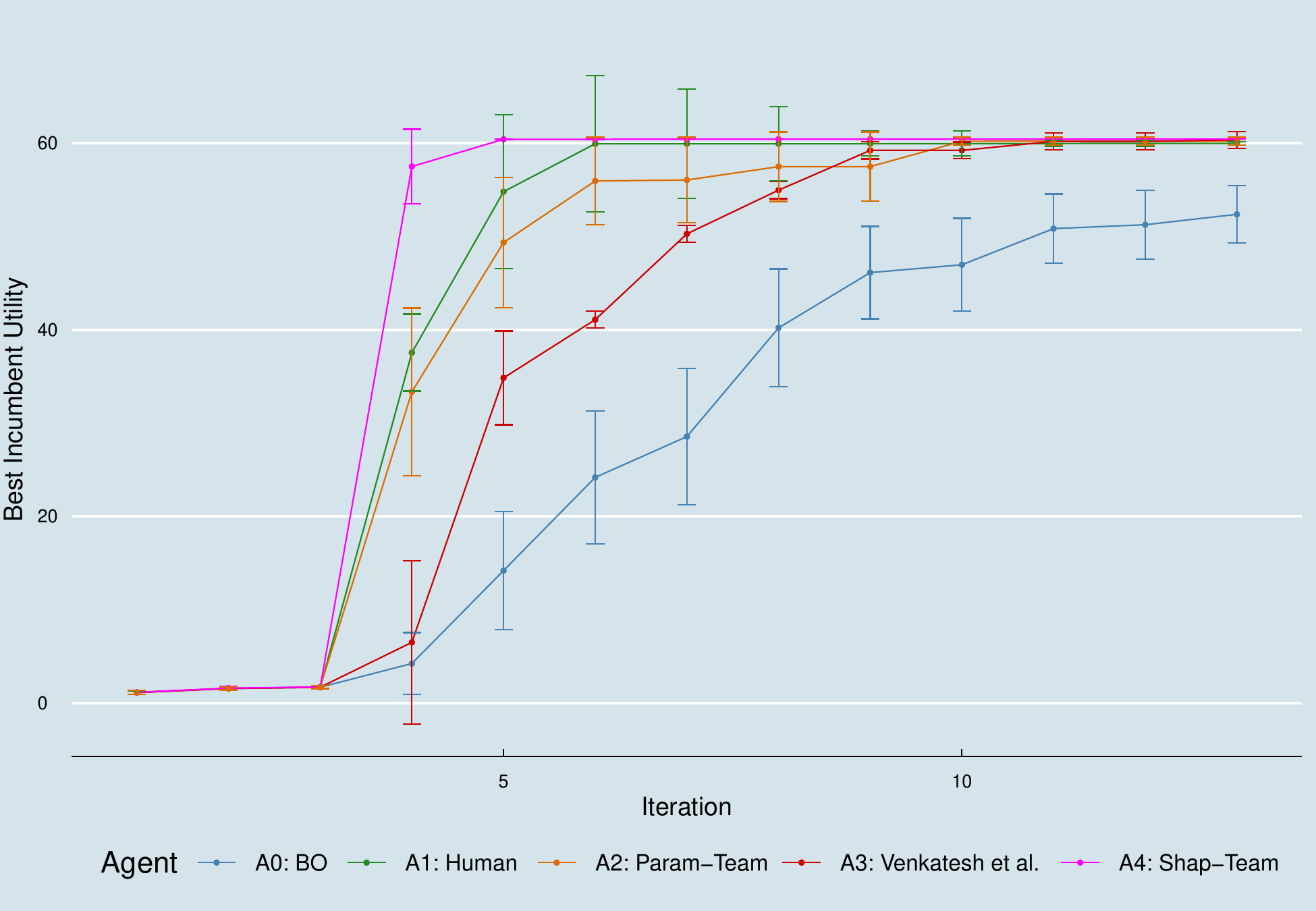}
    \caption{Results of Agents A0-A4 (see Table~\ref{tab:my_label}) in human-AI collaborative BO for simulated exosuit personalization (individual 1) with 10 iterations and 3 initial samples each. Error bars indicate $95 \%$ confidence intervals; $k=2$ for A3, $\beta = 2$ for A2 and A4.}
    \label{fig:results}
\end{figure}

We simulate 40 personalization rounds with 10 iterations and initial design of size 3 each for all five agents. We compare them with respect to \textit{optimization paths}, which show best incumbent target values (utility) in a given iteration. The BO uses GP as AM and cb with $\lambda = 20$ as AF; the BO modelling human proposals uses GP and cb with $\lambda = 200$ and prior knowledge of $90$ data points.     
%
%
%
%

Figure~\ref{fig:results} exemplarily summarizes results for individual~1; results for remaining 14 individuals as well as experimental details can be found in the supplement. For all 15 subjects, \texttt{ShapleyBO}-assisted A4 (Shap-team) outperforms human and BO baseline as well as \cite{human-bo-team} and a team that bases their decision to intervene on the proposed parameters. This latter comparison particularly confirms that Shapley values are a meaningful measure for human-BO alignment that cannot be replaced by another notion of alignment without loss of efficiency. For 10 (among whom is subject 1, see Figure~\ref{fig:results}) out of 15 subjects the observed outperformance is significant at $95\%$ confidence level.


\section{Discussion}

By quantifying the contribution of each parameter to the proposals, \texttt{ShapleyBO} aids in the communication of the rationale behind specific optimization decisions. This interpretability is not only crucial for trust in HIL applications, it also enhances their efficiency. 
The use case of customizing exosuits illustrates the practical benefits of this approach, suggesting that ShapleyBO could be a valuable practical tool for personalizing soft back exosuits. 

Our paper opens up a multitude of directions for future work. First and foremost, the simulation results in Section~\ref{sec:application} based on real-world data motivate the actual deployment of Shapley-assisted human-in-the-loop optimization of exosuits in a user study. On the methodological end, extensions of \texttt{ShapleyBO} to multi-criteria BO appear straightforward, as long as additive acquisition functions are used. Moreover, a thorough mathematical study of the sublinear regret bounds of Bayesian optimization with confidence bound \cite{srinivas2012information} under Shapley-assisted human interventions might foster theoretical understanding of why Shapley-assisted teams outperform competitors.

\section*{Ethical Statement}
The data used to estimate the utility functions was collected as part of another study \cite{Arens2023Preference} for which ethical approval was obtained from the applicable institutional ethical review board.
CJW is an inventor of at least one patent application describing the exosuit components described in the paper that have been filed with the U.S. Patent Office by Harvard University. Harvard University has entered into a licensing agreement with Verve Inc., in which CJW has an equity interest and a board position. The other authors declare that they have no competing interests.

\section*{Acknowledgments}

We acknowledge support by the Federal Statistical Office of Germany within the co-operation project \say{Machine Learning in Official Statistics}. JR further acknowledges support by the Bavarian Academy of Sciences (BAS) through the Bavarian Institute for Digital Transformation (bidt) and by the LMU mentoring program of the Faculty for Mathematics, Informatics, and Statistics. YS is supported by the DAAD program Konrad Zuse Schools of Excellence in Artificial Intelligence, sponsored by the Federal Ministry of Education and Research.

\newpage\null\thispagestyle{empty}\newpage

\appendix

\section{Supplement}

\subsection{Details on Experiments}

Recall from the main paper that in our experimental setup, humans have the capability to modify the Bayesian optimization process by adjusting the algorithm's suggestions, as illustrated in the pseudo-code below, which we print here again for ease of readability. We evaluate the performance of our \texttt{ShapleyBO}-enhanced human-BO collaboration against the approach described in \cite[Algorithm 1]{human-bo-team} and three other benchmark groups (solely human, solely BO, and a human-BO pair with an alternate intervention mechanism). The modeling of human decisions employs an additional layer of Bayesian optimization, inspired by \cite{borji2013bayesian,human-bo-team}, where $\bm \theta^{human} = (\theta^{human}_{lif},\theta^{human}_{low})^T$ is derived through an acquisition function that captures human preferences. This auxiliary BO uses the same surrogate and acquisition models as the primary loop but adjusts for human-specific risk tolerance and initial knowledge through different exploration-exploitation preference ($\lambda$ in $cb$) and initial designs, respectively. The specifications for both the original BO and the auxiliary BO modeling the human can be found below.

\subsubsection{Model Specifications}\label{sec:setup}
The following non-exhaustive list summarizes the specifications for the conducted experiments.

\begin{itemize}
    \item Bayesian Optimization
        \begin{itemize}
            \item Surrogate Model: Gaussian Process
            \begin{itemize}
                \item Kernel: Gaussian
                $$k(x,x') \exp(-1/2 \cdot ( \lvert x - x' \rvert /\theta)^2),$$
                where $\theta$ is the range parameter which was estimated from data (empirical Bayes)
                \item Prior mean function estimated from data (empirical Bayes)
            \end{itemize}
                \item Acquisition Function: Confidence Bound
                    \begin{itemize}
                        \item $\lambda_{bo} = 20$
                    \end{itemize}
                \item initial design size: $3$
        \end{itemize}
    \item Human
            \begin{itemize}
            \item Surrogate Model: Gaussian Process
            \begin{itemize}
                \item Kernel: Gaussian
                $$k(x,x') \exp(-1/2 \cdot ( \lvert x - x' \rvert /\theta)^2),$$
                where $\theta$ is the range parameter which was estimated from data (empirical Bayes)
                \item Prior mean function estimated from data (empirical Bayes)
            \end{itemize}
                            \item Acquisition Function: Confidence Bound
                    \begin{itemize}
                        \item $\lambda_h = 200$
                    \end{itemize}
                \item initial design size: $90$ (from a subset of the feature space, modeling prior knowledge of the human)
        \end{itemize}
\end{itemize}

\setcounter{algorithm}{0}
\begin{algorithm}

\begin{center}
\begin{small}    
\caption{Human-AI Collaborative BO}
\label{algo-bo_human}

    \begin{algorithmic}[1]
    \State create an initial design $D = \{(\bm{\theta}^{(i)}, \Psi^{(i)})\}_{i = 1,..., n_{init}}$ 
    \While {termination criterion is not fulfilled}
    \State \textbf{train} SM on data $D$
    \State \textbf{propose} $\bm{\theta}^{new}$ that optimizes $AF(SM(\bm{\theta}))$ 
    \If {intervention criterion is fulfilled} \label{alg:intervent-crit}
    \State $\bm{\theta}^{new} \leftarrow \bm{\theta}^{human}$ 
    \EndIf
    \State \textbf{evaluate} $\Psi$ on $\bm{\theta}^{new}$ 
    \State \textbf{update} $D \leftarrow D \cup {(\bm{\theta}^{new}, \Psi(\bm{\theta}^{new}))}$ 
    \EndWhile
    \State \textbf{return} $\argmin_{\bm{\theta} \in D} \Psi(\bm{\theta})$ 
    \end{algorithmic}
\end{small}
    \end{center}
\end{algorithm}

\begin{table}[b]
  \scalebox{.6}{
    \centering
    \begin{tabular}{c|ccccc}
        Agent & A0 & A1 & A2 & A3 & A4  \\
         & BO & Human & Param-Team & \cite{human-bo-team} & Shap-Team \\
         \hline
        Intervention &&&&& \\
        Criterion & never  & always & $\theta^{new}_{lif},\theta^{new}_{low}$ & $k$-th iteration & $\phi_{lif}^{new}(m), \phi_{low}^{new}(m)$

    \end{tabular}
    }
    \caption{All agents: \texttt{ShapleyBO}-assisted A4 and baselines A0-A3. }
    \label{app-tab:my_label}
\end{table}

Varying both $\lambda_h$ and $\lambda_{bo}$ did not affect the overall results significantly. Drastically decreasing initial design size for the human led to all human-BO teams (A2, A3, A4) to be outperformed by the the BO Baseline A0. We reason that a certain advantage in knowledge by the humans is key to successful human-BO collaboration.

\subsubsection{Details on Simulations}

We computed 40 repetitions of the described procedure on a Linux High-performance cluster. The experiments were tested with \texttt{R} versions 4.3.2, 4.2.0, 4.1.6, and 4.0.3 on Linux Ubuntu 20.04, Linux Debian 10, and Windows 11 Pro Build 22H2.  
Further details as well as documentation of how to reproduce the results can be found in the readme-file of the anonymous repository:\\ 

\url{https://anonymous.4open.science/r/ShapleyBO-E65D}

\subsubsection{Alternative Intervention Criteria}\label{sec:tree}

\begin{figure}[H]
    \centering
    \includegraphics[width=1\linewidth]{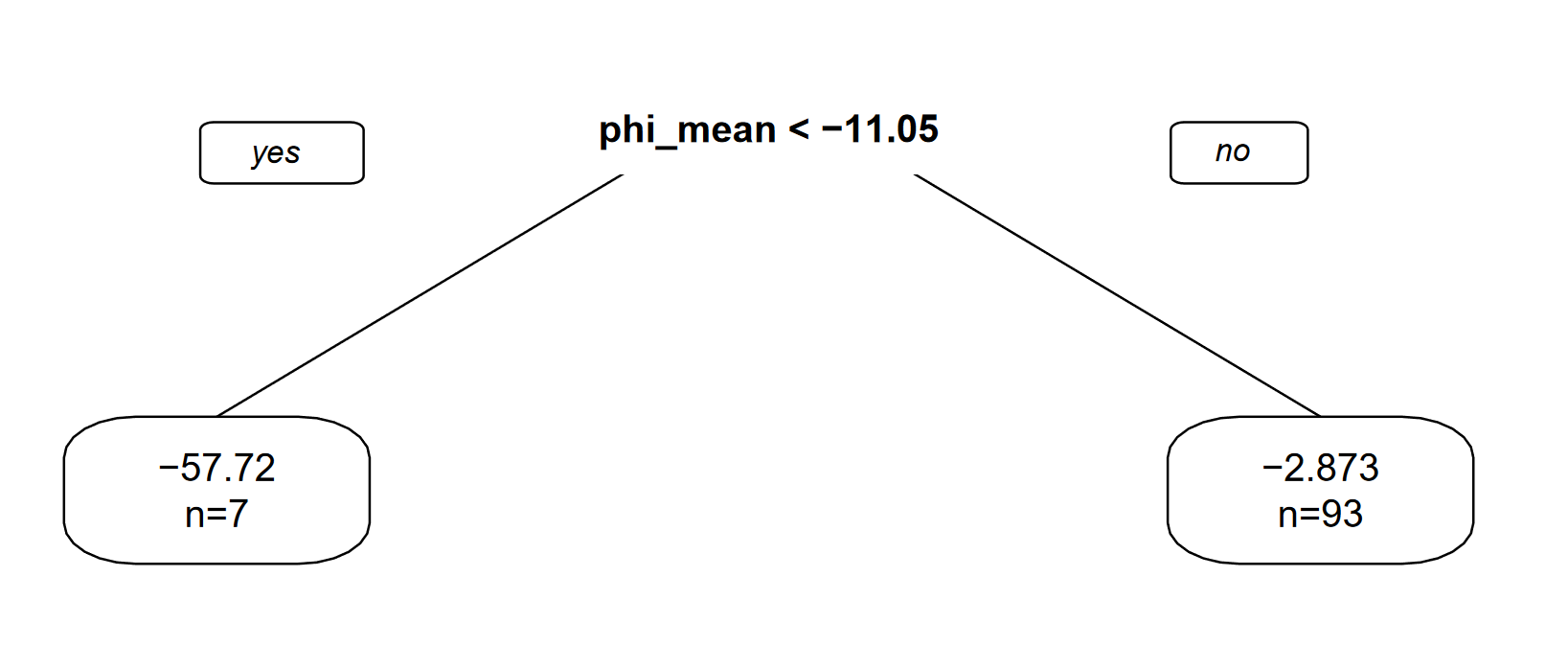}
    \caption{Regression tree learned from Shapley-assisted BO. It predicts a target value of $\Psi(\bm \theta) = -57.72$ for BO proposals where $\theta_{lif}$ (lifting gain) has a Shapley contribution to the mean component of cb lower than $-11.05$, i.e., $\phi_{lif}(m) < -11.05$. Using this learned rule as an intervention criterion in human-AI collaborative BO translates to rejecting the automatically generated BO proposals and rectifying them by human's proposals in case their respective Shapley values fulfill $\phi_{lif}(m) < -11.05$.}
    \label{app-fig:tree}
\end{figure}

We varied $\beta$ for both A2 and A4 and did not observe significant changes in the results. Moreover, we considered different intervention criteria based on previous experience of the human rather than hard coded like Equations 11 and 12 in the main paper. The idea was to learn the criterion from previous BO runs. The so acquired data (proposals $\bm \theta$ and respective Shapley values as well as the evaluated target $\Psi(\bm \theta)$) was used to train an ML model to represent the learned knowledge in previous BO iteration. To this end, we rely on classification and regression trees (CART), as they are easy to interpret, flexible and robust \cite{loh2014fifty,fernandez2014we,nalenz2024learning}.
We particularly rely on their ability to learn logical rules that can then be used as intervention criteria. The pruned version of a tree delivers a decision rule of required complexity, see Figure~\ref{app-fig:tree}. It represents the human's knowledge about the relationship between the proposals' Shapley values and the obtained target values.

\begin{figure}[H]
    \centering
    \includegraphics[width=1\linewidth]{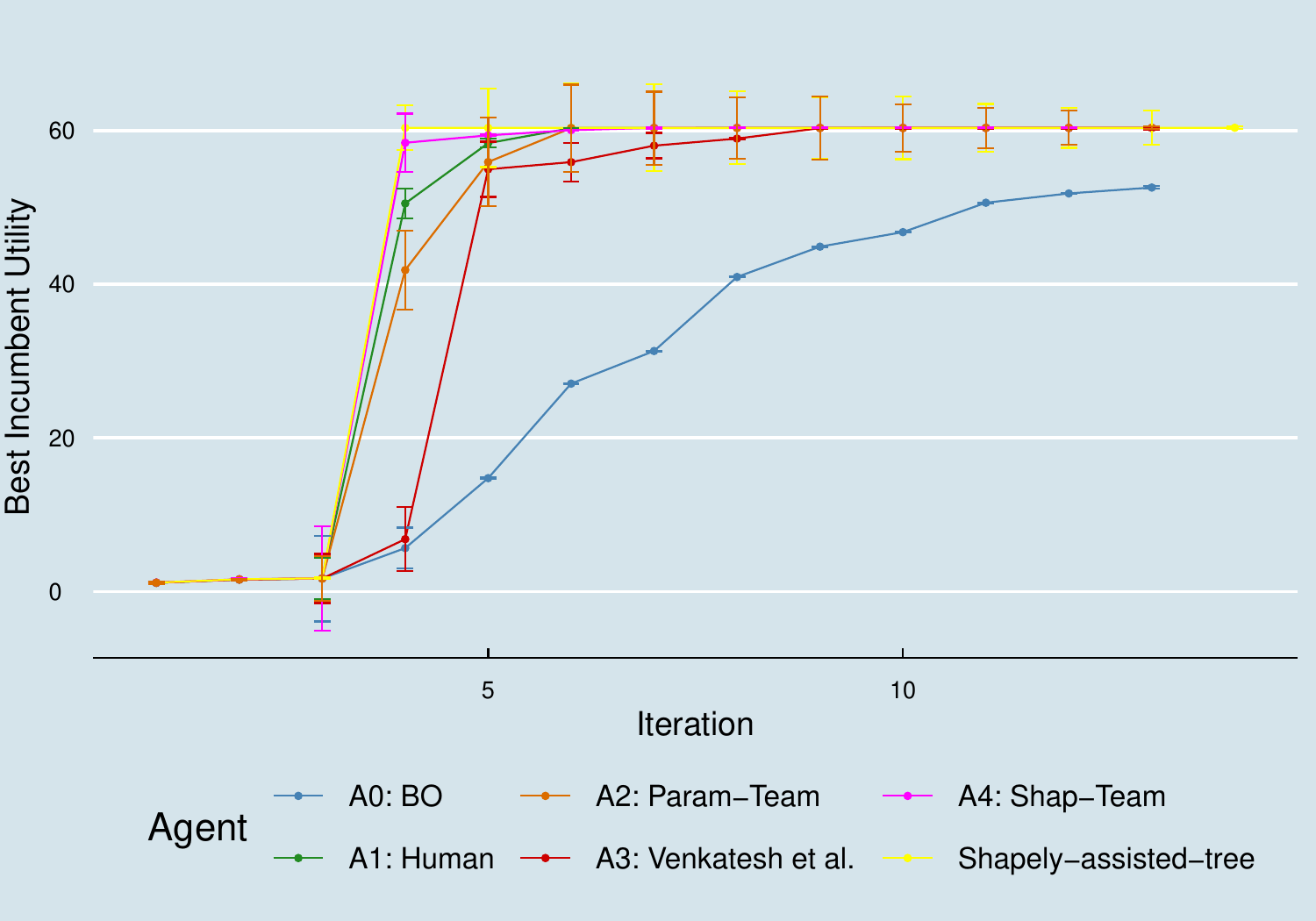}
    \caption{Results of Agents A0-A4 (see Table~\ref{app-tab:my_label}) and additional agent with tree-based intervention criterion in human-AI collaborative BO for simulated exosuit personalization (individual 1) with 10 iterations and 3 initial samples each. Error bars indicate $95 \%$ confidence intervals; $k=2$ for A3, $\beta = 2$ for A2 and A4.}
    \label{app-fig:tree-results}
\end{figure}

\begin{figure}[H]
    \centering
    \includegraphics[width=1\linewidth]{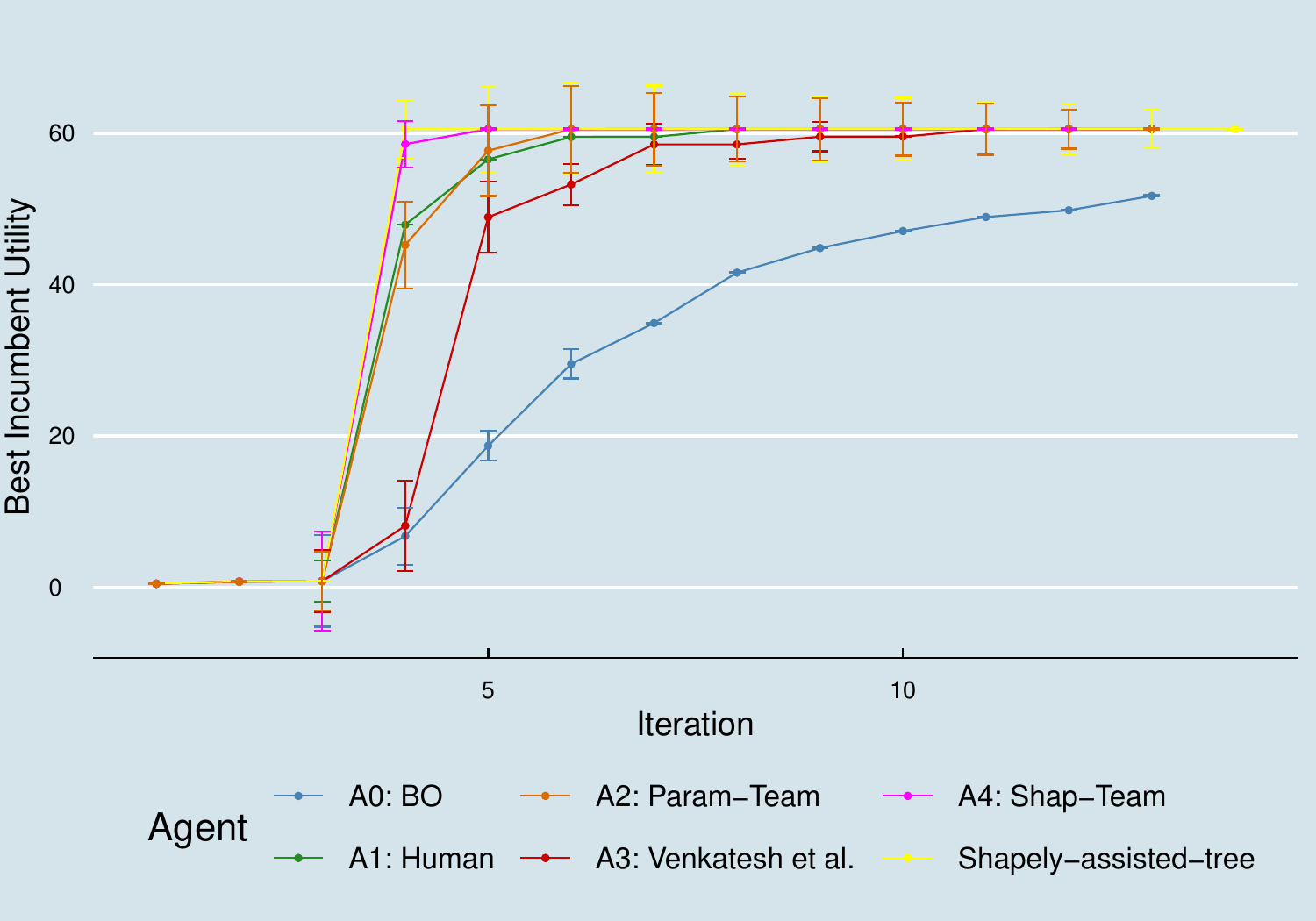}
    \caption{Results of Agents A0-A4 (see Table~\ref{app-tab:my_label}) and additional agent with tree-based intervention criterion in human-AI collaborative BO for simulated exosuit personalization (randomly selected individual) with 10 iterations and 3 initial samples each. Error bars indicate $95 \%$ confidence intervals; $k=2$ for A3, $\beta = 2$ for A2 and A4.}
    \label{app-fig:tree-results-2}
\end{figure}

Replacing the hard-coded intervention criteria (Equation~11 and~12 in the main paper) by those learned through a regression tree did not change the results significantly, see Figure~\ref{app-fig:tree-results} and Figure~\ref{app-fig:tree-results-2}. We ran the same experiments as described in the main paper, but additionally included a Shapley-assisted human-BO team with intervention criterion learned through regression trees (yellow).

As becomes evident, the reported results are relatively robust with regard to the concrete specification of the intervention criterion. That is, the precise nature of the criterion (see also discussion of results for altered $\beta$) and whether it is hard-coded or learned from previous BO iterations does not alter the results significantly.

\subsubsection{Research Hypotheses}

Experimental design was motivated by the following hypotheses that were formulated beforehand.

\begin{Hypothesis}[BO Baseline]
\label{hypo:interventions-good}
Human-BO teams are more efficient than BO alone.
\end{Hypothesis}

\begin{Hypothesis}[Human Baseline]
\label{hypo:bo-good}
    Human-BO teams are more efficient than a human without access to BO.
\end{Hypothesis}

\begin{Hypothesis}[\cite{human-bo-team} Baseline]
\label{hypo:shapleys-better}
An agent in human-AI collaborative BO that intervenes based on Shapley values is more efficient than an agent who does intervene in each $k$-th iteration.
\end{Hypothesis}

\begin{Hypothesis}[Parameter-informed Baseline]
\label{hypo:shapleys-better}
An agent in human-AI collaborative BO that intervenes based on Shapley values is more efficient than an agent who intervenes based on the discrepancy between the proposed parameter values.
\end{Hypothesis}


Efficiency in Hypotheses~\ref{hypo:interventions-good} through~\ref{hypo:shapleys-better} refers to the cumulative regret (distances to optimum summed over all iterations).
Hypotheses~\ref{hypo:interventions-good} and~\ref{hypo:bo-good} make sure Shapley-assisted human-BO teams do not simply outperform competitors due to the fact that the human is uniformly superior to BO, or vice versa. Notably, hypotheses~\ref{hypo:interventions-good} through~\ref{hypo:shapleys-better} were treated as scientific research hypotheses rather than statistical hypotheses. We leave a rigorous statistical benchmarking analysis along the lines of \cite{demvsar2006statistical,jansen2023statistical} to future work as well as multi-criteria comparison \cite{pmlr-v216-jansen23a,pmlr-v215-blocher23a,rodemann2024partial} of the optimization agents.


\subsection{Additional Experiments and Future Work}

\begin{figure}[H]
    \centering
    \includegraphics[width=1\linewidth]{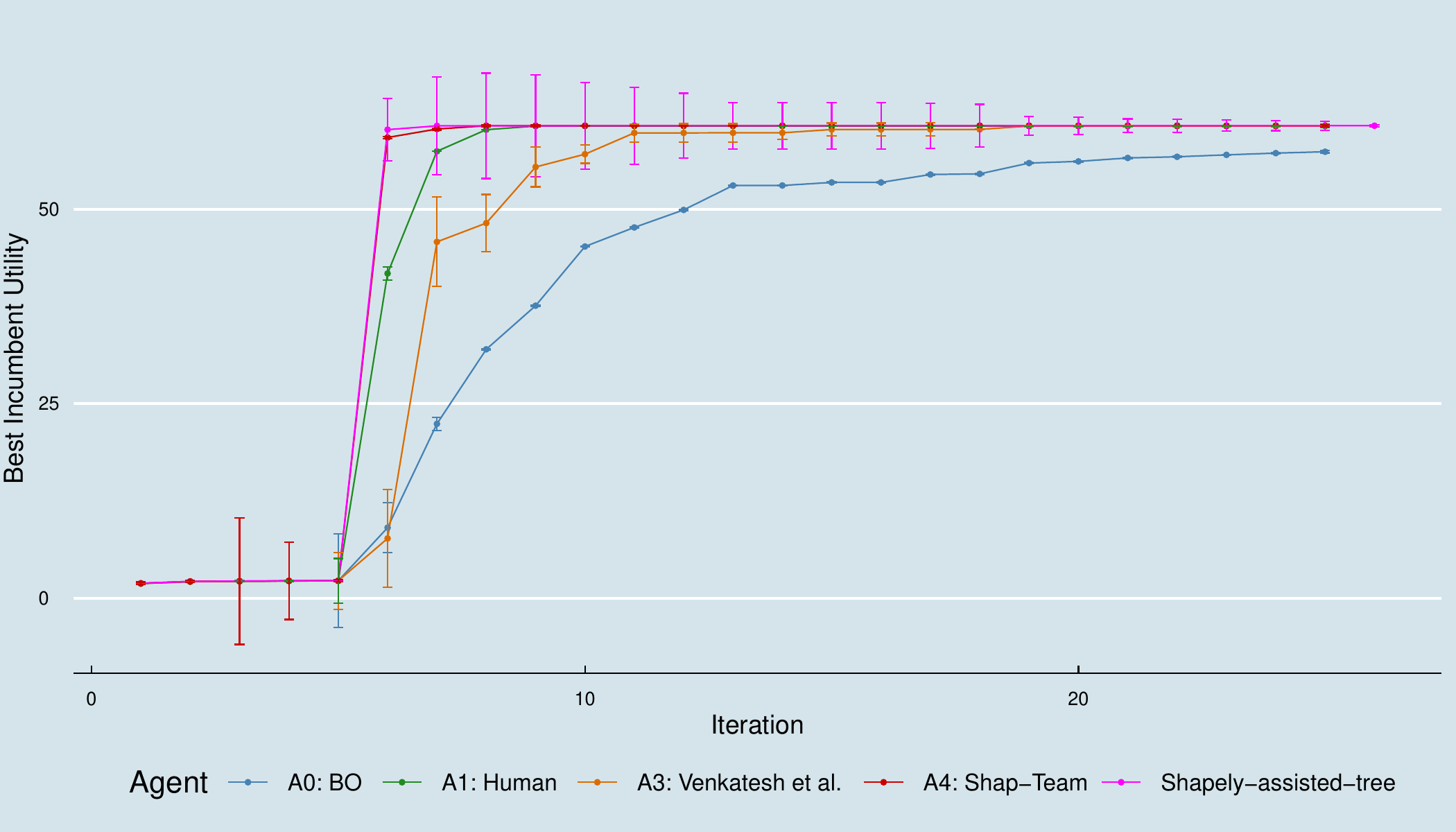}
    \caption{Results of for simulated exosuit personalization (randomly selected individual) for $\lambda_h = \lambda_{bo} = 40$ with 20 iterations and 5 initial samples each. Error bars indicate $95 \%$ confidence intervals.}
    \label{app-fig:vary-lambda}
\end{figure}

As already hinted at in the subsections on model specifications and alternative intervention criteria in Section~\ref{sec:tree}, we did conduct some robustness analysis. The results from varying the intervention criteria can be see in Figures~\ref{app-fig:tree-results} and~\ref{app-fig:tree-results-2} and have been discussed above. In what follows, we exemplarily showcase some results from experiments where we altered $\lambda_h$ and $\lambda_{BO}$. Figure~\ref{app-fig:vary-lambda} has the results for $\lambda_h = \lambda_{bo} = 40$, i.e. a more balanced exploration-exploitation preference between human and BO. The results are hardly affected by the more balanced exploration-exploitation scheme.

Besides the above presented additional experiments for exosuit personalization, we also tested \texttt{ShapleyBO} in a univariate scenario, where deployment of the uncertainty-aware confidence bound (Equation 7 in main paper) was possible. We focused on the concrete case of minimizing metabolic cost by step size adaption of powered prosthetics for walking assistance \cite{kim2017human}. The publicly available data contains measurements on 9 patients. We use each of them to fit a surrogate function that we optimize by BO. The proposal are interpreted by Shapley values, showcasing the potential use of disentangling the uncertainty contributions even in a univariate setup.

\subsection{Computational details}

A principal hurdle in applying Shapley values (SVs) within the realm of interpretable machine learning (IML) is their computational intensity. To precisely calculate Shapley values, one must evaluate the model across all conceivable feature subsets. Such exact calculations are known to possess exponential time complexity \cite{vstrumbelj2014explaining}, making the use of approximation methods such as Monte Carlo (MC) sampling a practical necessity. 

For instance, \cite{merrick2019} devised the Explanation Game, an innovative sampling approach that employs importance sampling to enhance the accuracy of Shapley value estimates.\cite{ignatov2022} introduced a method rooted in formal concept analysis for assessing Shapley values over finite attribute sets, while \cite{pmlr-v206-luther23a} applied causal structure learning to pinpoint feature independencies, thereby streamlining Shapley value computation.

In our work, we adopt a traditional MC sampling strategy, supplemented by a novel algorithm from \cite{croppi2021explaining}, designed to accurately determine an adequate sample size for Shapley value estimation in Bayesian optimization contexts. This algorithm presents a viable means of identifying the requisite sample volume for efficient SV computation. Let us define the payout function

\begin{equation}
\mathcal{P}(v) = \hat{f}^{v}(\tilde{\bm{\theta}}) - \frac{1}{n}\sum_{i}\hat{f}^{v}(\bm{\theta}^{(i)}),    
\end{equation}

intended for distribution among the explicand's parameters $\tilde{\bm{\theta}}$, where $\hat{f}^{v}$ predicts outcomes $v = \{cb, m, se\}$ and $\{\bm{\theta}^{(i)}\}_{i =1}^{n}$ denotes a series of instances. As per the \textbf{efficiency axiom}, the equality $\sum_{j}\phi_{j}(v) = \mathcal{P}(v)$ is expected. However, due to approximation in calculating true contributions, this process incurs an \textit{efficiency error}, represented as 

\begin{equation}
\Delta_{eff}^{K}(v) = \sum_{j}\hat{\phi}_{j}^{K}(v) - \mathcal{P}(v),
\end{equation}

with $K$ symbolizing the MC sample count. The discrepancy $\Delta_{eff}^{K}(v)$ diminishes as $K$ increases, aligning with the asymptotic distribution of $\hat{\phi}_{j}(v)$. An error greater (lesser) than zero indicates an overallocation (underallocation) of payout, necessitating an adjustment in contributions. 
While the precise contributions requiring adjustment remain uncertain, redistributing $\Delta_{eff}^{K}(v)$ entirely to a single parameter would arguably be the least equitable.

Our focus lies not on rectifying the outcomes but on evaluating the adequacy of $K$. For computational simplicity, we redefine $\Delta_{eff}^{K}(v)$ as $|\sum_{j}\hat{\phi}_{j}^{K}(v) - \mathcal{P}(v)|$.
Moreover, we introduce 

\begin{equation}
\delta^{K}(v) = min\{\bm{d_{1}}(\hat{\phi}_j^{K}(v), \hat{\phi}_l^{K}(v))_{j \neq l}\},
\end{equation}

the minimal $L_1$ distance between any two distinct parameter Shapley values. A sample size $K$ is deemed sufficient if $\Delta_{eff}^{K}(v) < \delta^{K}(v)$, ensuring that adjustments for efficiency error do not alter the importance rankings. If this condition is not met, $K$ must be increased.

A sufficient sample size is thus determined through a greedy forward search. Below, a visual representation of this methodology is provided, and Algorithm \ref{algo:check_ss} concisely outlines the primary procedural steps, see also the respetive pseudo-code in the following section.

\begin{figure}[H]
    \centering
    \includegraphics[scale = 0.35]{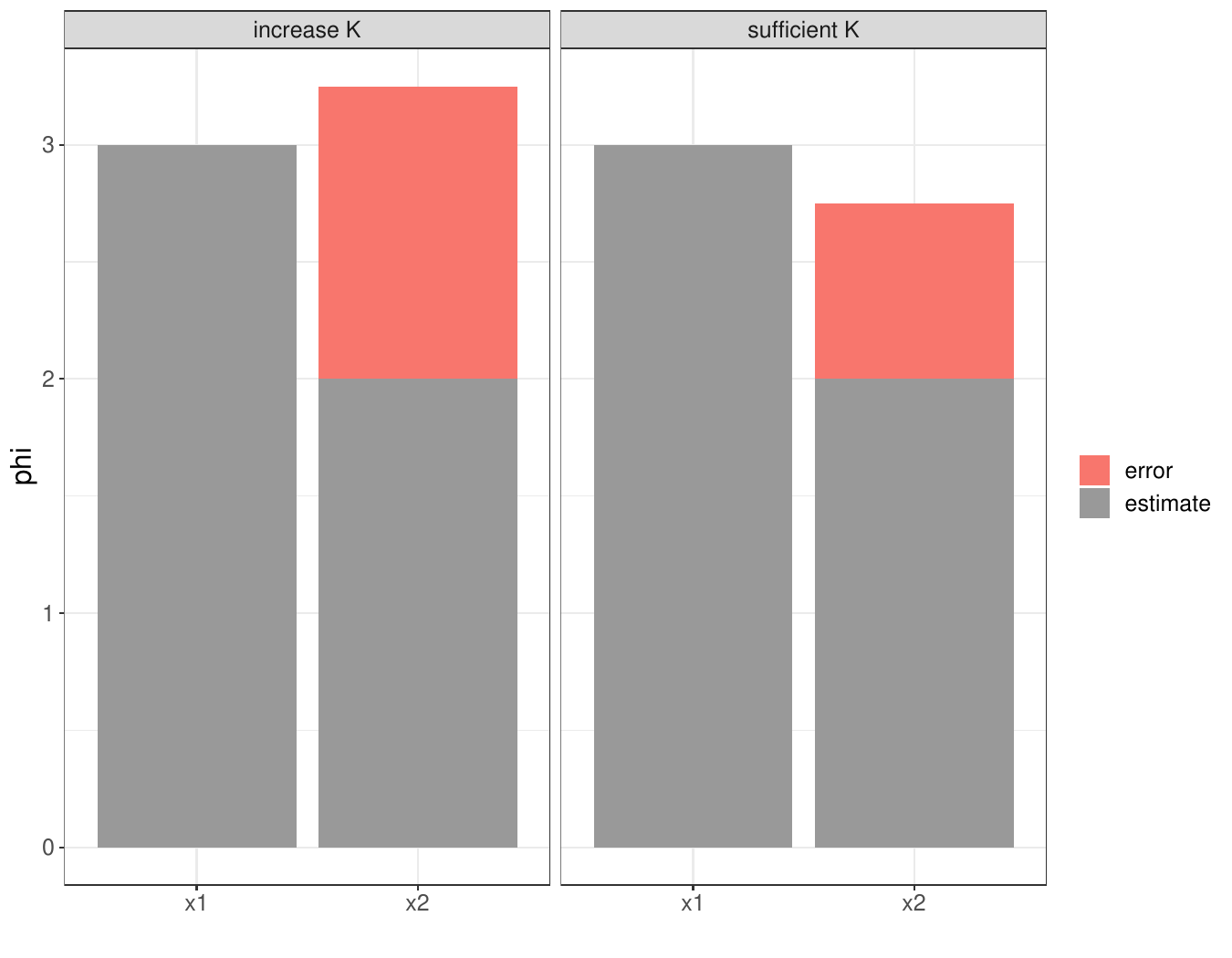}
    \caption[Example of \texttt{checkSampleSize}]{Example of the idea behind \texttt{checkSampleSize}, see \cite[Figure 16]{croppi2021explaining}. Estimated Shapley values of two features named $x_1$ and $x_2$ in grey give threshold $\delta^K =3-2= 1$. The efficiency error in red varies for different $K$. Suppose $\Delta_{eff}^{K}(v) < 0$. The error is thus added entirely to $x_2$. In the left plot the sample size $K$ is not sufficient because redistributing the resulting error would make $x_2$ more important than $x_1$. Instead, in the right plot $K$ would be sufficient. In case of $\Delta_{eff}^{K}(v) > 0$ removing the error completely from $x_1$ would lead to the same conclusion for $K$.} 
    \label{app-fig:check_ss}
\end{figure}

\subsection{Pseudo-Code}

In what follows, we summarize the main algorithmic procedures in pseudo-code: The general estimation of the Shapley values, the main procedure to retrieve Shapley values from any iteration in Bayesian Optimization, and the above mentioned procedure to check the sample size required for MC-estimation of the Shapley values. Detailed implementation of all procedures in \texttt{R} \cite{R2020} (dependencies: \texttt{mlrMBO} \cite{bischl2017mlrmbo}, \texttt{iml} \cite{molnar2018iml}) can be accessed in folder R of:\\

\url{https://anonymous.4open.science/r/ShapleyBO-E65D}

\begin{algorithm}[H]
\begin{center}
\caption{Estimation of the Shapley values}
\label{algo:sv_ces}
    \begin{algorithmic}[1]
    \Require explicand $\tilde{\bm{\theta}}$, feature index $j$, model $\hat{f}$ and sample size $K$ 
    \For{$k = 1 \to K$}
        \State sample (with replacement) an instance $\bm{z} \in \bm{\Theta}$
        \State sample (with replacement) an order $\pi \in \Pi(P)$
        \State order $\tilde{\bm{\theta}}$ and $\bm{z}$ according to $\pi$
        \State \hspace{0.5cm}$\tilde{\bm{\theta}}_{\pi} = (\tilde{\theta}_{(1)},\ldots,\tilde{\theta}_{(p)})$
        \State \hspace{0.5cm}$\bm{z}_{\pi} = (z_{(1)},\ldots,z_{(p)})$
        \State construct two new instances
        \State \hspace{0.5cm}$\tilde{\bm{\theta}}_{+j} = (\tilde{\theta}_{(1)},\ldots, \tilde{\theta}_{(j-1)}, \tilde{\theta}_{(j)}, z_{(j+1)}, \ldots,  z_{(p)})$
        \State \hspace{0.5cm}$\tilde{\bm{\theta}}_{-j} = (\tilde{\theta}_{(1)},\ldots, \tilde{\theta}_{(j-1)}, z_{(j)}, z_{(j+1)}, \ldots,  z_{(p)})$
        \State $\hat{\phi}_j^k(v) = \hat{f}(\tilde{\bm{\theta}}_{+j}) - \hat{f}(\tilde{\bm{\theta}}_{-j})$
    \EndFor
    \State $\hat{\phi}_j(v) = \frac{1}{K}\sum_{k=1}^{K}\hat{\phi}_{j}^{k}(v)$
    \end{algorithmic}
    \end{center}
\end{algorithm}

\begin{algorithm}[H]
\caption{\texttt{ShapleyBO}}
\label{algo:shapley_bo}
  \begin{algorithmic}[1]
    \Require BO result object \textit{bo}, iteration of interest $t$, sample size $K$
    \State get explicand from \textit{bo}: $\tilde{\bm{\theta}} = \bm{\theta}^{new}_t $
    \State sample $1000\cdot p$ points $\bm{Z}$ from $\bm{\Theta}$ to approximate the space
    \State compute $\hat{\bm{\phi}}(m) = (\hat{\phi}_{1}(m), \ldots, \hat{\phi}_{p}(m))$:
        \State \hspace{0.5cm} get SM from \textit{bo}: $\hat{f}_{m} =  \hat{f}_{t}^{mean} $
        \State \hspace{0.5cm} explain $\tilde{\bm{\theta}}$ with \texttt{Shapley()} using $\bm{Z}$, $\hat{f}_m$ and $K$
    \State  compute $\hat{\bm{\phi}}(se) = (\hat{\phi}_{1}(se), \ldots, \hat{\phi}_{p}(se))$:
        \State \hspace{0.5cm} get SM from \textit{bo}: $\hat{f}_{se} =  \hat{f}_{t}^{uncertainty} $
        \State \hspace{0.5cm} explain $\tilde{\bm{\theta}}$ with \texttt{Shapley()} using $\bm{Z}$, $\hat{f}_{se}$ and $K$
    \State compute $\hat{\bm{\phi}}(cb)$ with linearity axiom: $\hat{\bm{\phi}}(cb) = \hat{\bm{\phi}}(m) - \lambda \hat{\bm{\phi}}(se)$
  \end{algorithmic}
\end{algorithm}

\begin{algorithm}[H]
\caption{\texttt{checkSampleSize}}
\label{algo:check_ss}
  \begin{algorithmic}[1]
    \Require \texttt{ShapleyBO} results for $\tilde{\bm{\theta}}_t$ with size $K$, models $\hat{f}^v$ with $v = \{cb, m, se\}$
\For {$w$ in $v$}
    \State compute payout $\mathcal{P}(w)$, error $\Delta_{eff}^{K}(w)$\; and threshold $\delta^{K}(w)$
    \If{$\Delta_{eff}^{K}(w) < \delta^{K}(w)$} 
        \State $K_w$ = \texttt{T}
    \EndIf
    \If{$\Delta_{eff}^{K}(w) \geq \delta^{K}(w)$} 
        \State $K_w$ = \texttt{F}
    \EndIf
\EndFor
\If{$(K_{cb},K_m,K_{se})=(\texttt{T}, \texttt{T}, \texttt{T})$}
    \State $K$ is high enough
\ElsIf{$(K_{cb},K_m,K_{se}) \neq (\texttt{T}, \texttt{T}, \texttt{T})$}
    \State $K$ should be increased
\EndIf    
  \end{algorithmic}
\end{algorithm}

\newpage
\subsection{Additional Results}
\begin{figure}[H]
    \centering
    \includegraphics[width=0.98\linewidth]{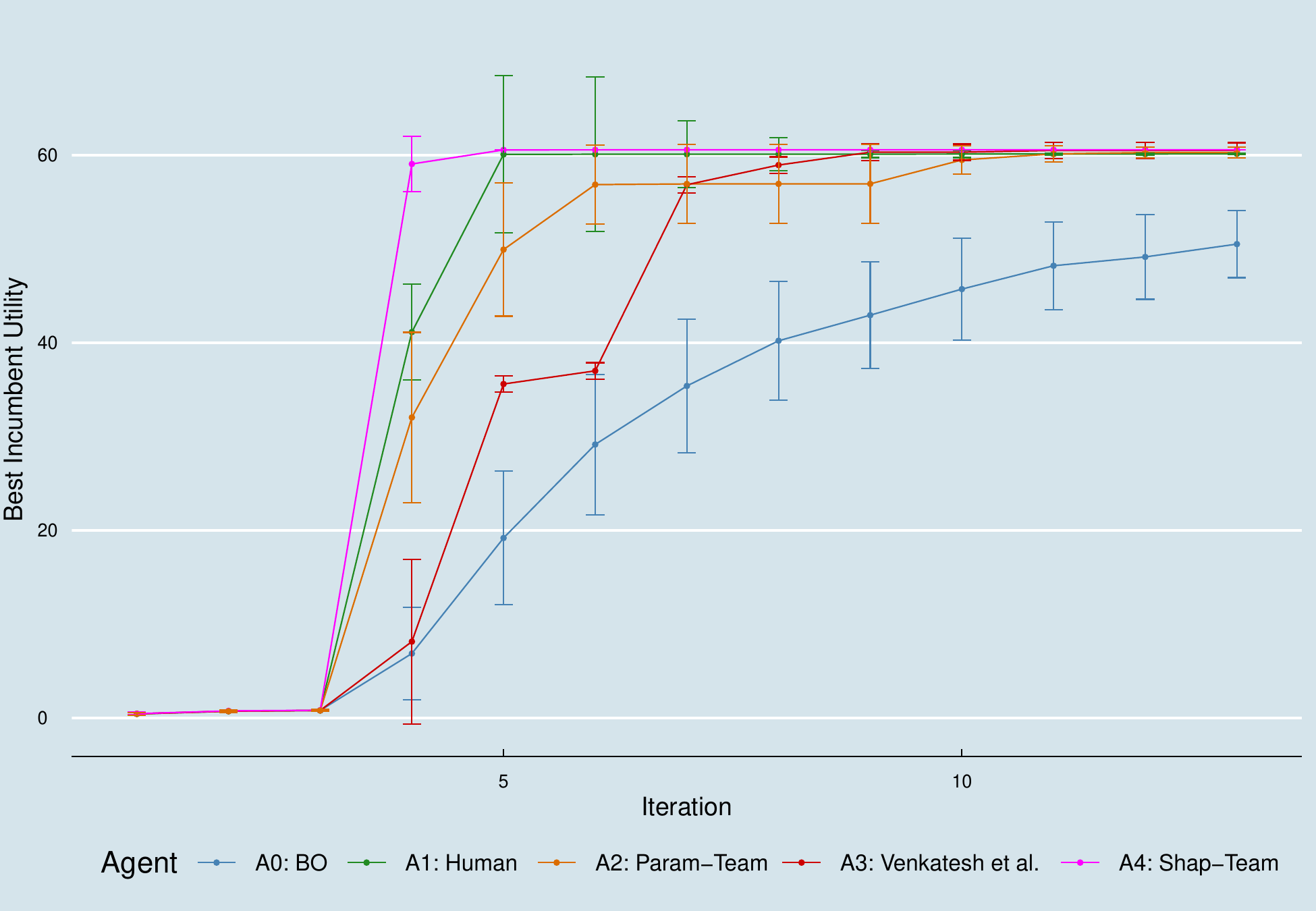}
    \caption{Results of Agents A0-A4 (see Table~\ref{app-tab:my_label}) in human-AI collaborative BO for simulated exosuit personalization (individual 2) with 10 iterations and 3 initial samples each. Error bars indicate $95 \%$ confidence intervals; $k=2$ for A3, $\beta = 2$ for A2 and A4. }
    \label{app-fig:S2}
\end{figure}

\begin{figure}[H]
    \centering
    \includegraphics[width=0.98\linewidth]{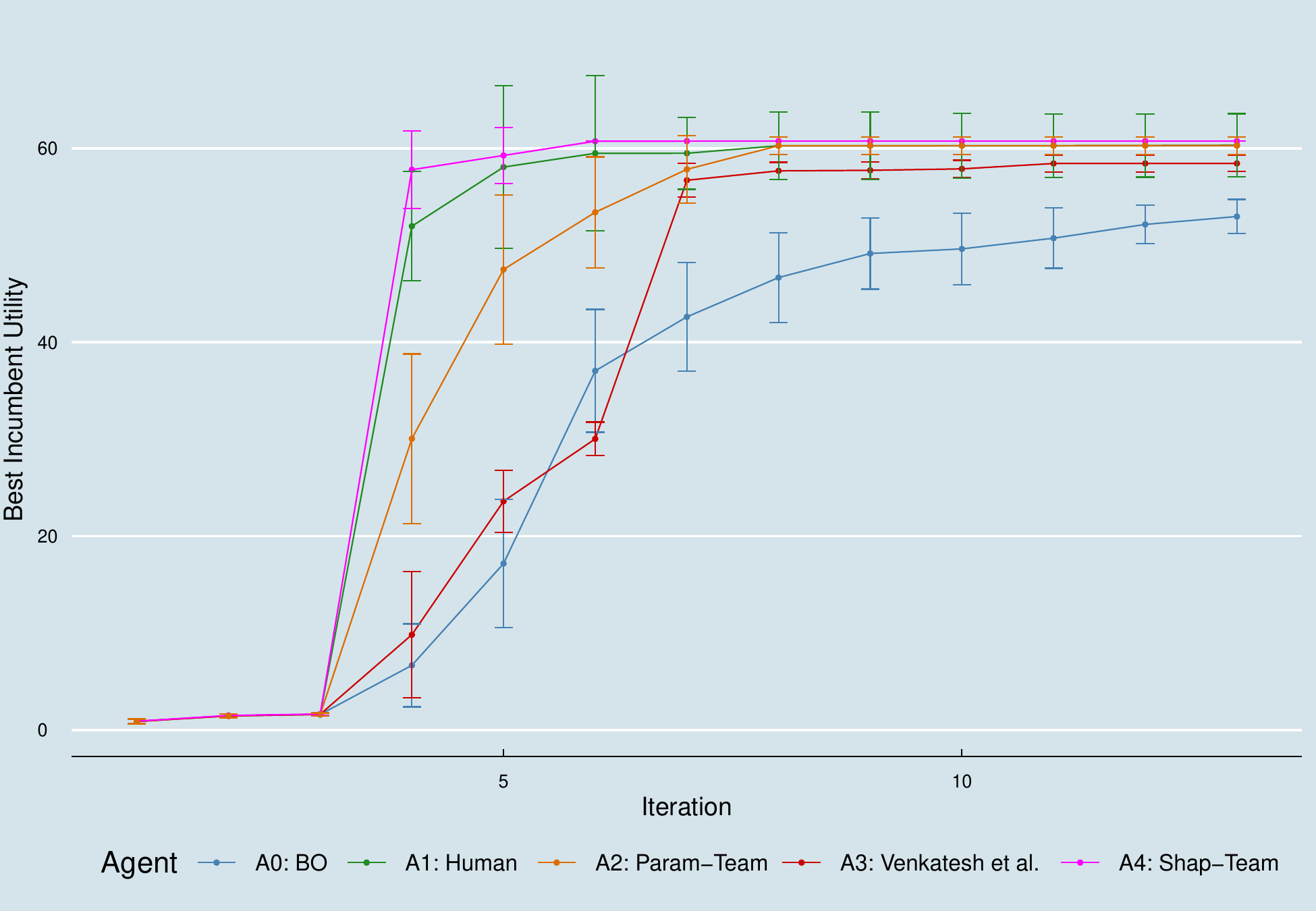}
    \caption{Results of Agents A0-A4 (see Table~\ref{app-tab:my_label}) in human-AI collaborative BO for simulated exosuit personalization (individual 3) with 10 iterations and 3 initial samples each. Error bars indicate $95 \%$ confidence intervals; $k=2$ for A3, $\beta = 2$ for A2 and A4.}
    \label{app-fig:S3}
\end{figure}

\begin{figure}[H]
    \centering
    \includegraphics[width=0.98\linewidth]{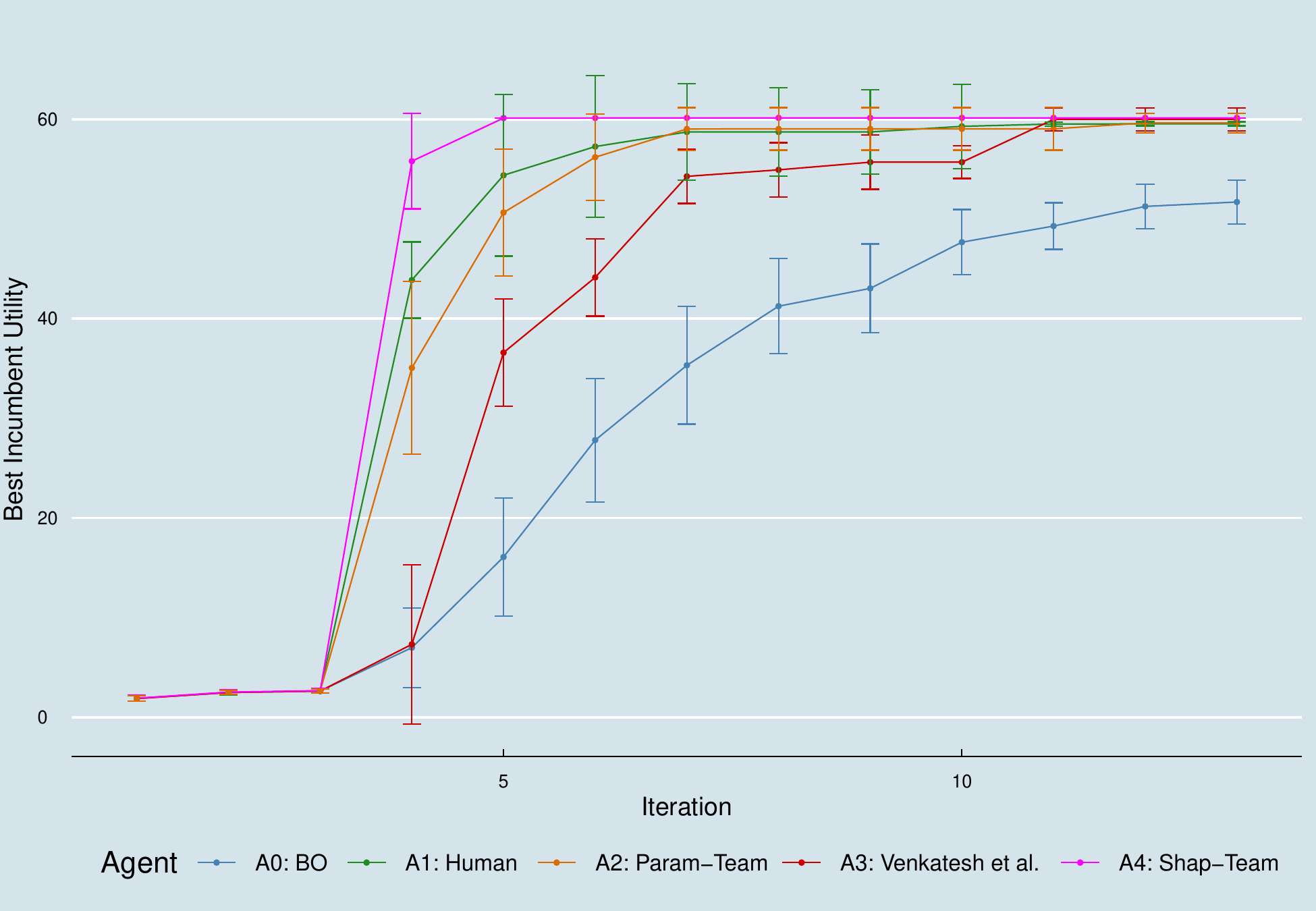}
    \caption{Results of Agents A0-A4 (see Table~\ref{app-tab:my_label}) in human-AI collaborative BO for simulated exosuit personalization (individual 4) with 10 iterations and 3 initial samples each. Error bars indicate $95 \%$ confidence intervals; $k=2$ for A3, $\beta = 2$ for A2 and A4.}
    \label{app-fig:S4}
\end{figure}

\begin{figure}[H]
    \centering
    \includegraphics[width=0.98\linewidth]{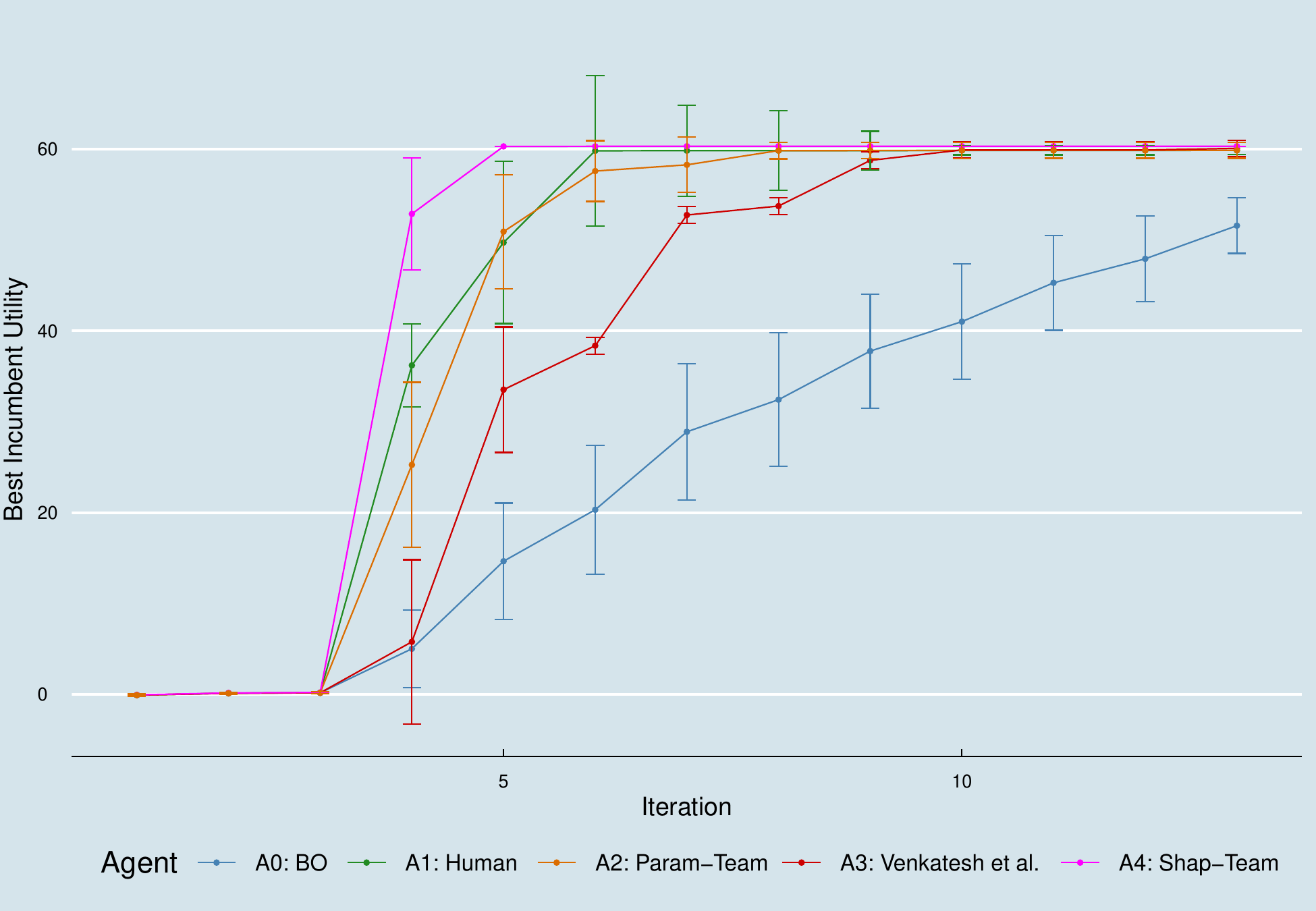}
    \caption{Results of Agents A0-A4 (see Table~\ref{app-tab:my_label}) in human-AI collaborative BO for simulated exosuit personalization (individual 5) with 10 iterations and 3 initial samples each. Error bars indicate $95 \%$ confidence intervals; $k=2$ for A3, $\beta = 2$ for A2 and A4.}
    \label{app-fig:S5}
\end{figure}

\begin{figure}[H]
    \centering
    \includegraphics[width=0.98\linewidth]{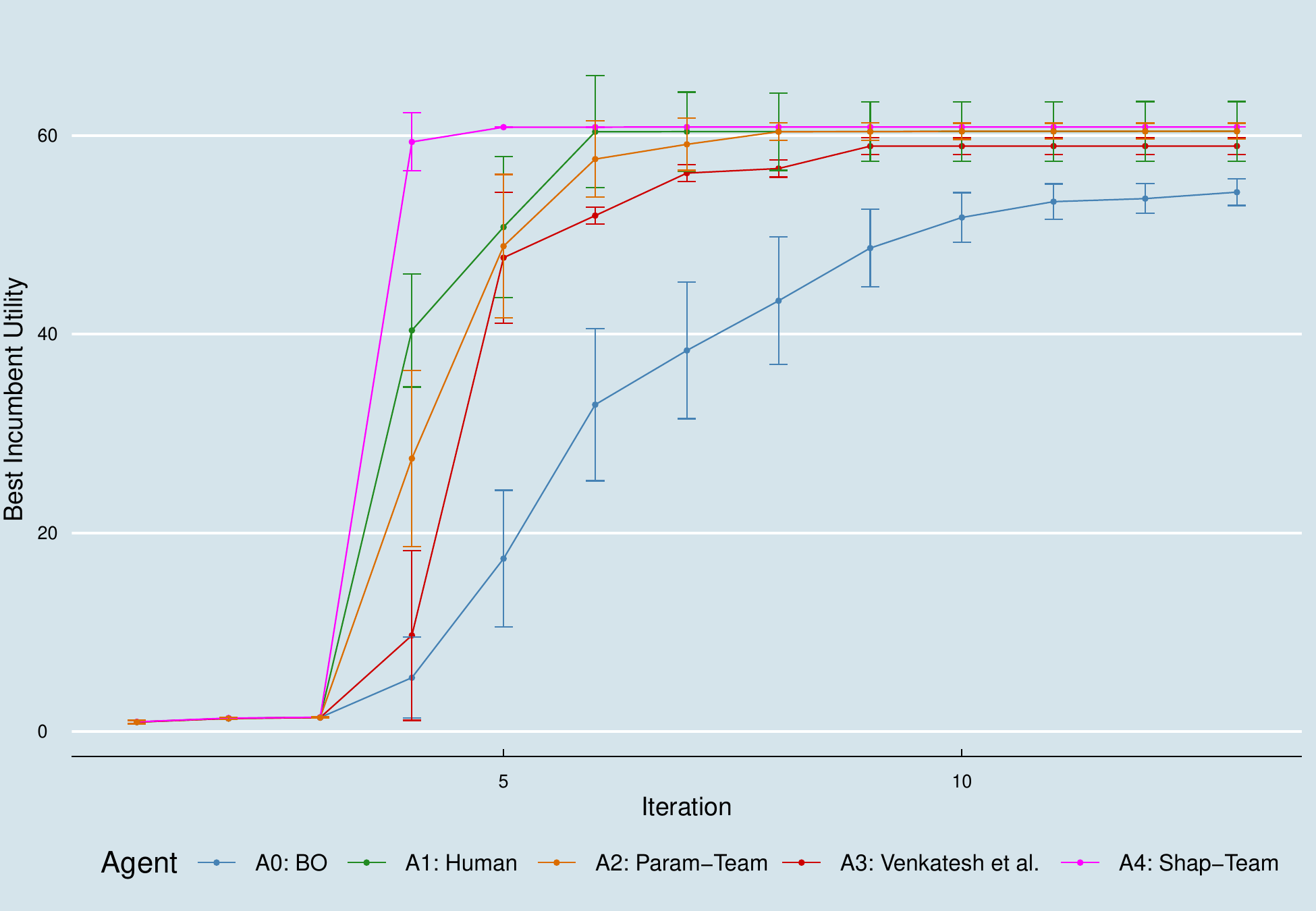}
    \caption{Results of Agents A0-A4 (see Table~\ref{app-tab:my_label}) in human-AI collaborative BO for simulated exosuit personalization (individual 6) with 10 iterations and 3 initial samples each. Error bars indicate $95 \%$ confidence intervals; $k=2$ for A3, $\beta = 2$ for A2 and A4.}
    \label{app-fig:S6}
\end{figure}

\begin{figure}[H]
    \centering
    \includegraphics[width=0.98\linewidth]{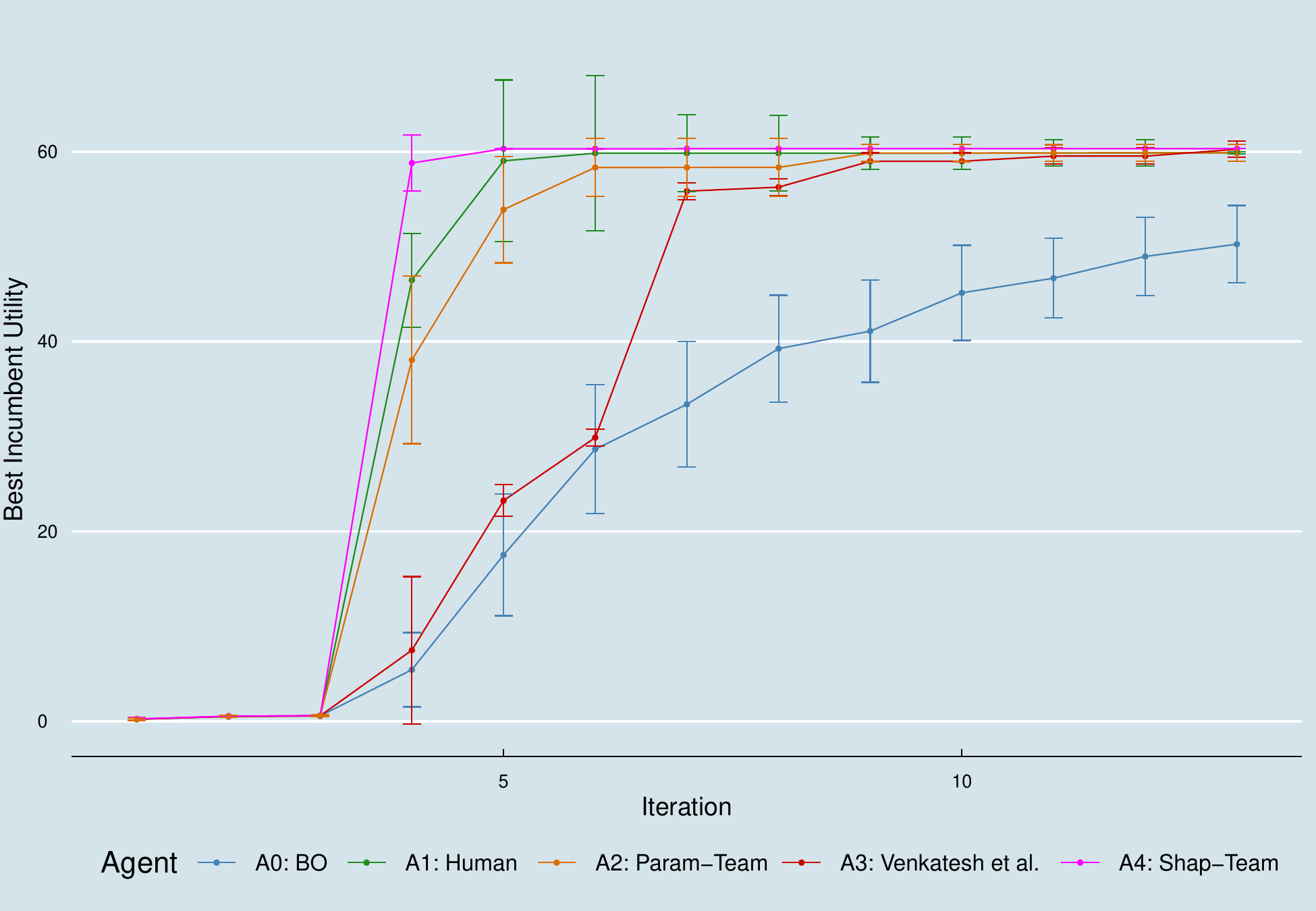}
    \caption{Results of Agents A0-A4 (see Table~\ref{app-tab:my_label}) in human-AI collaborative BO for simulated exosuit personalization (individual 7) with 10 iterations and 3 initial samples each. Error bars indicate $95 \%$ confidence intervals; $k=2$ for A3, $\beta = 2$ for A2 and A4.}
    \label{app-fig:S7}
\end{figure}

\begin{figure}[H]
    \centering
    \includegraphics[width=0.98\linewidth]{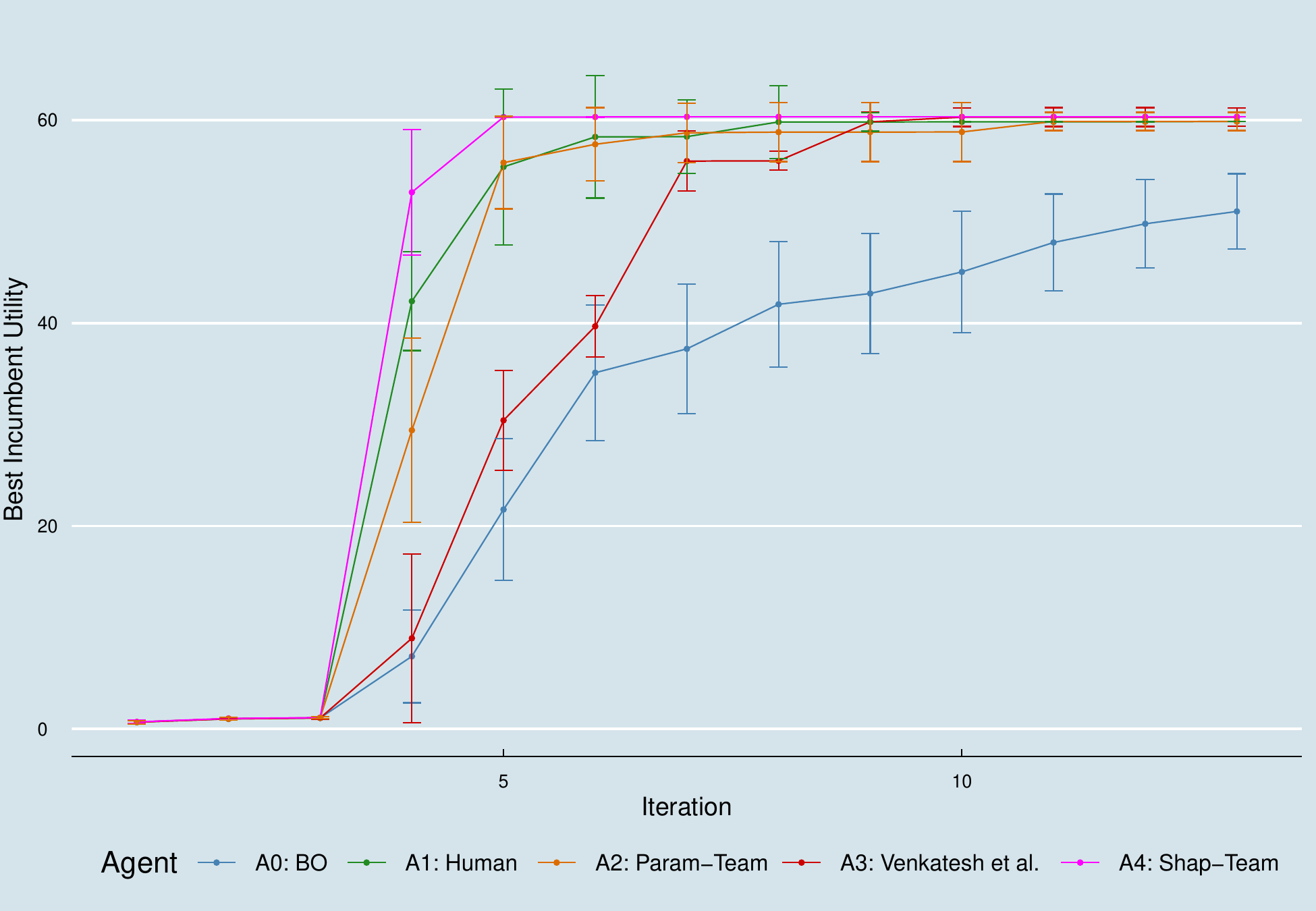}
    \caption{Results of Agents A0-A4 (see Table~\ref{app-tab:my_label}) in human-AI collaborative BO for simulated exosuit personalization (individual 8) with 10 iterations and 3 initial samples each. Error bars indicate $95 \%$ confidence intervals; $k=2$ for A3, $\beta = 2$ for A2 and A4.}
    \label{app-fig:S8}
\end{figure}

\begin{figure}[H]
    \centering
    \includegraphics[width=0.98\linewidth]{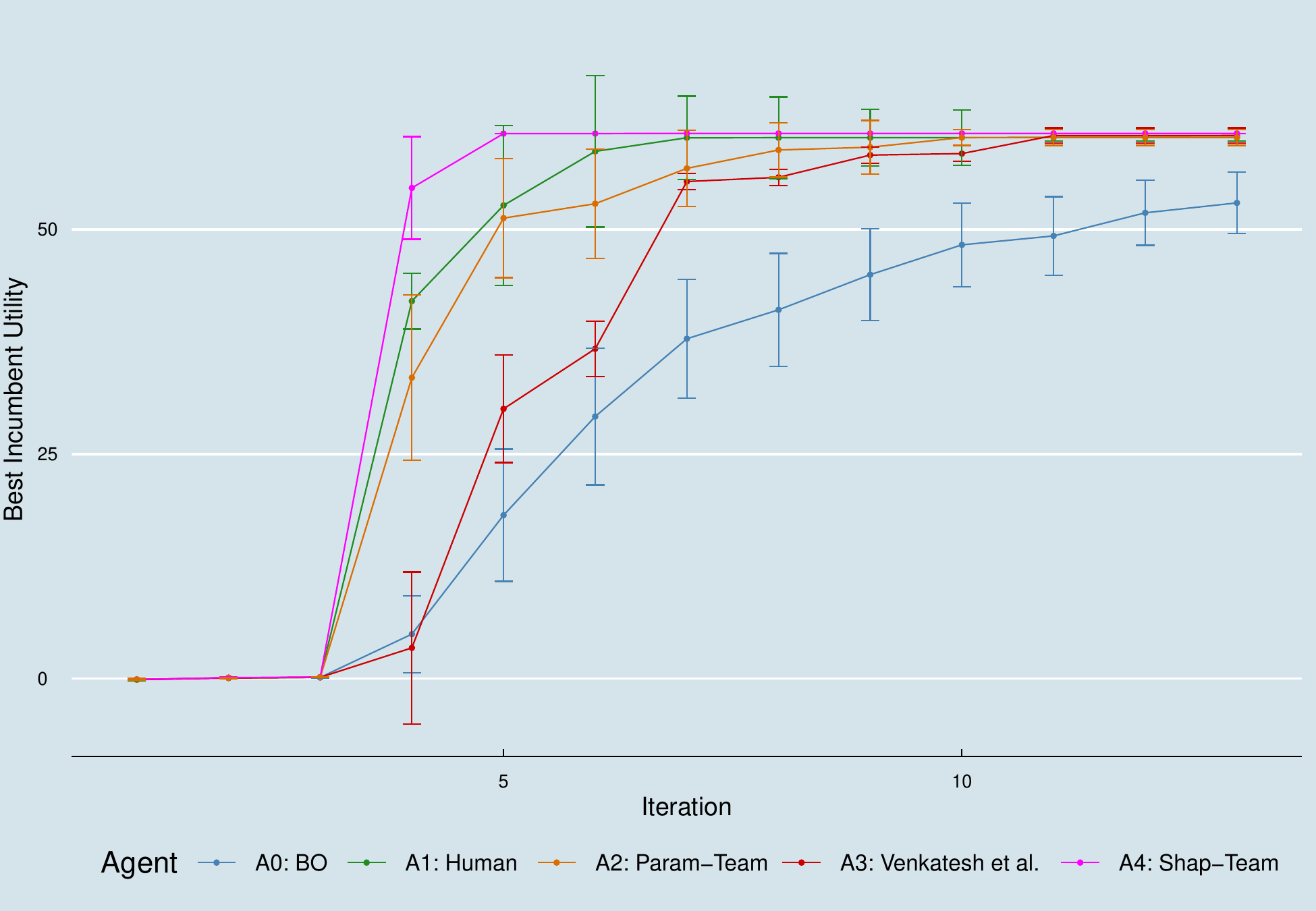}
    \caption{Results of Agents A0-A4 (see Table~\ref{app-tab:my_label}) in human-AI collaborative BO for simulated exosuit personalization (individual 9) with 10 iterations and 3 initial samples each. Error bars indicate $95 \%$ confidence intervals; $k=2$ for A3, $\beta = 2$ for A2 and A4.}
    \label{app-fig:S9}
\end{figure}

\begin{figure}[H]
    \centering
    \includegraphics[width=0.98\linewidth]{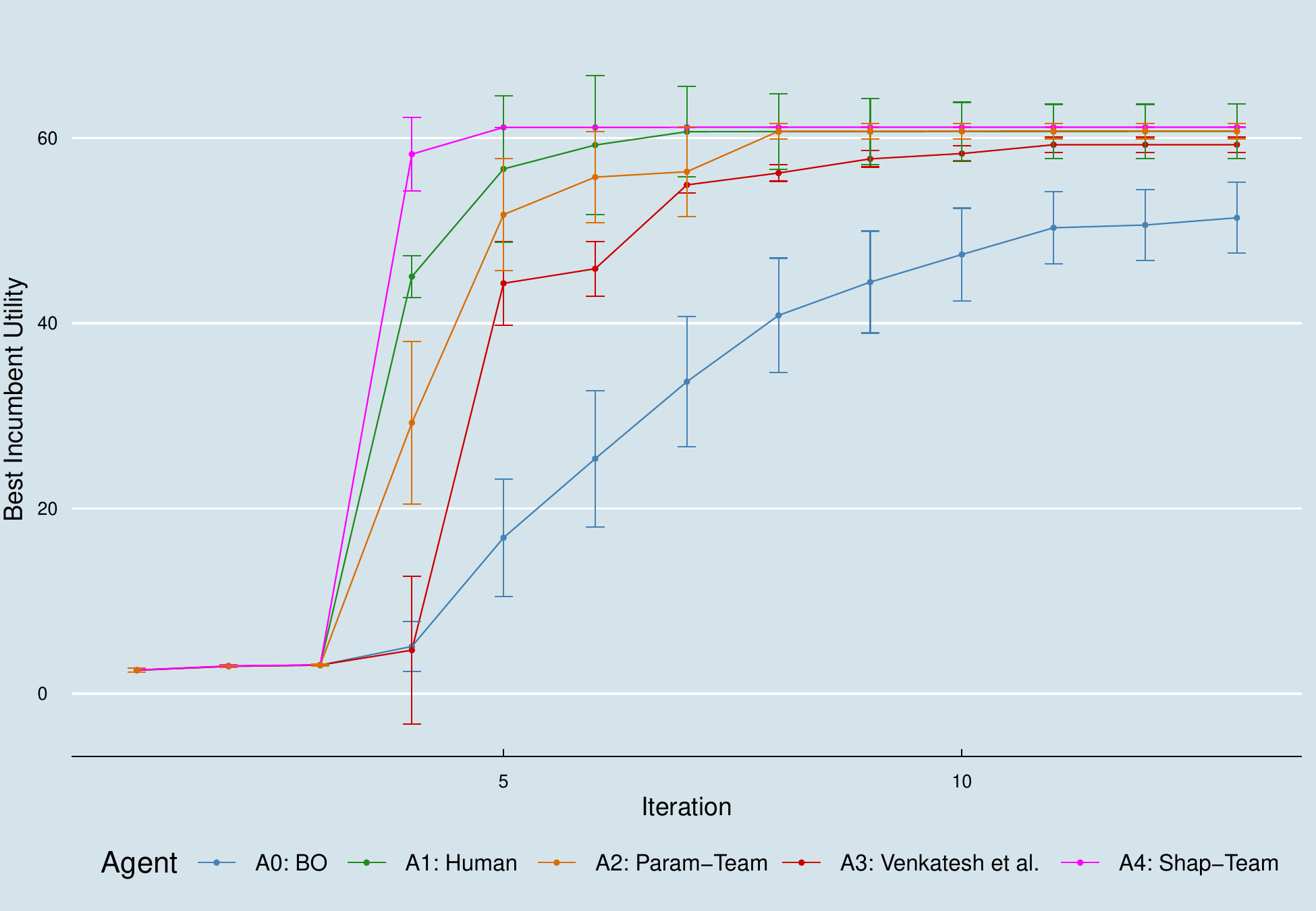}
    \caption{Results of Agents A0-A4 (see Table~\ref{app-tab:my_label}) in human-AI collaborative BO for simulated exosuit personalization (individual 10) with 10 iterations and 3 initial samples each. Error bars indicate $95 \%$ confidence intervals; $k=2$ for A3, $\beta = 2$ for A2 and A4.}
    \label{app-fig:S10}
\end{figure}

\begin{figure}[H]
    \centering
    \includegraphics[width=0.98\linewidth]{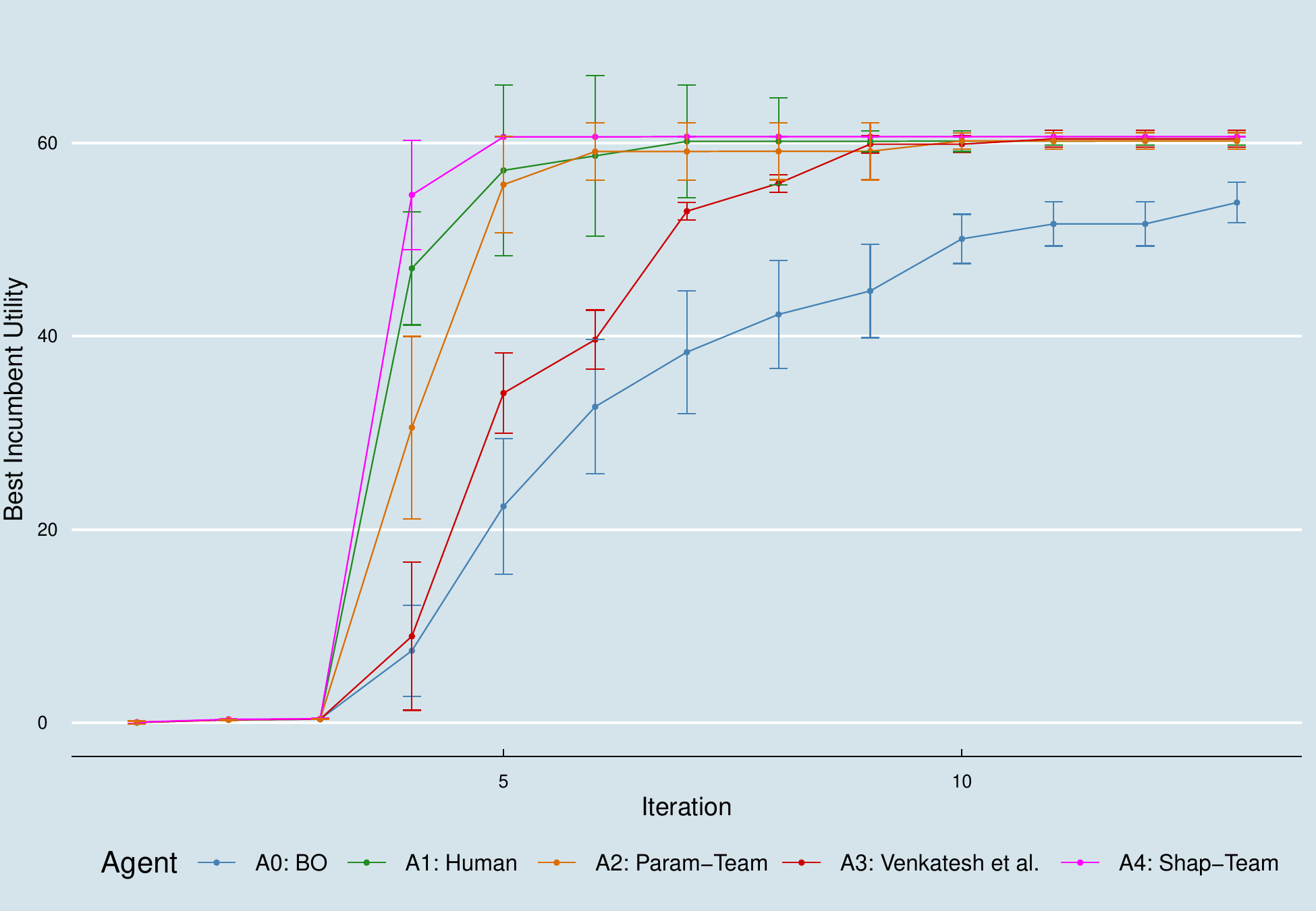}
    \caption{Results of Agents A0-A4 (see Table~\ref{app-tab:my_label}) in human-AI collaborative BO for simulated exosuit personalization (individual 11) with 10 iterations and 3 initial samples each. Error bars indicate $95 \%$ confidence intervals; $k=2$ for A3, $\beta = 2$ for A2 and A4.}
    \label{app-fig:S11}
\end{figure}

\begin{figure}[H]
    \centering
    \includegraphics[width=0.98\linewidth]{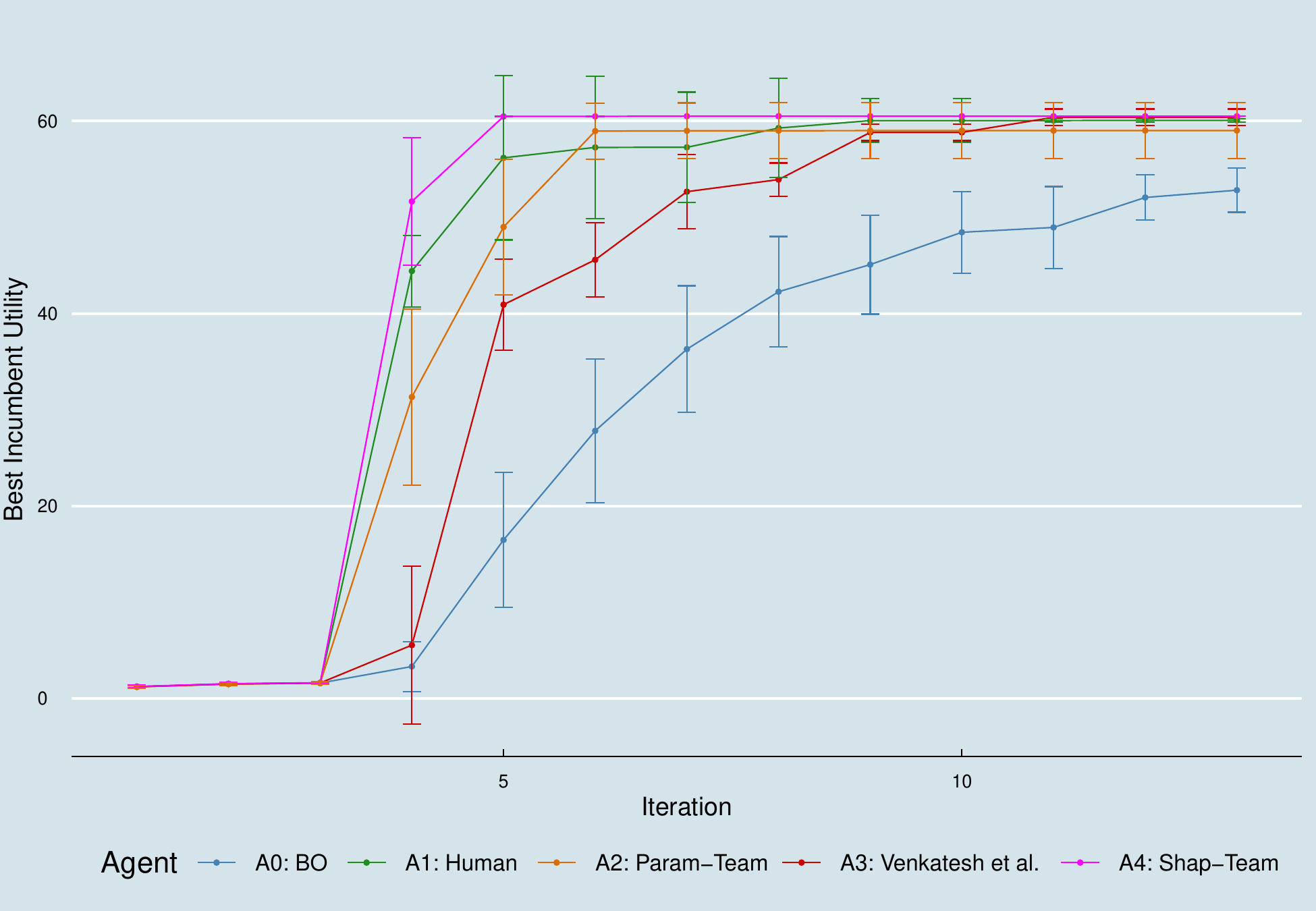}
    \caption{Results of Agents A0-A4 (see Table~\ref{app-tab:my_label}) in human-AI collaborative BO for simulated exosuit personalization (individual 12) with 10 iterations and 3 initial samples each. Error bars indicate $95 \%$ confidence intervals; $k=2$ for A3, $\beta = 2$ for A2 and A4.}
    \label{app-fig:S12}
\end{figure}

\begin{figure}[H]
    \centering
    \includegraphics[width=0.98\linewidth]{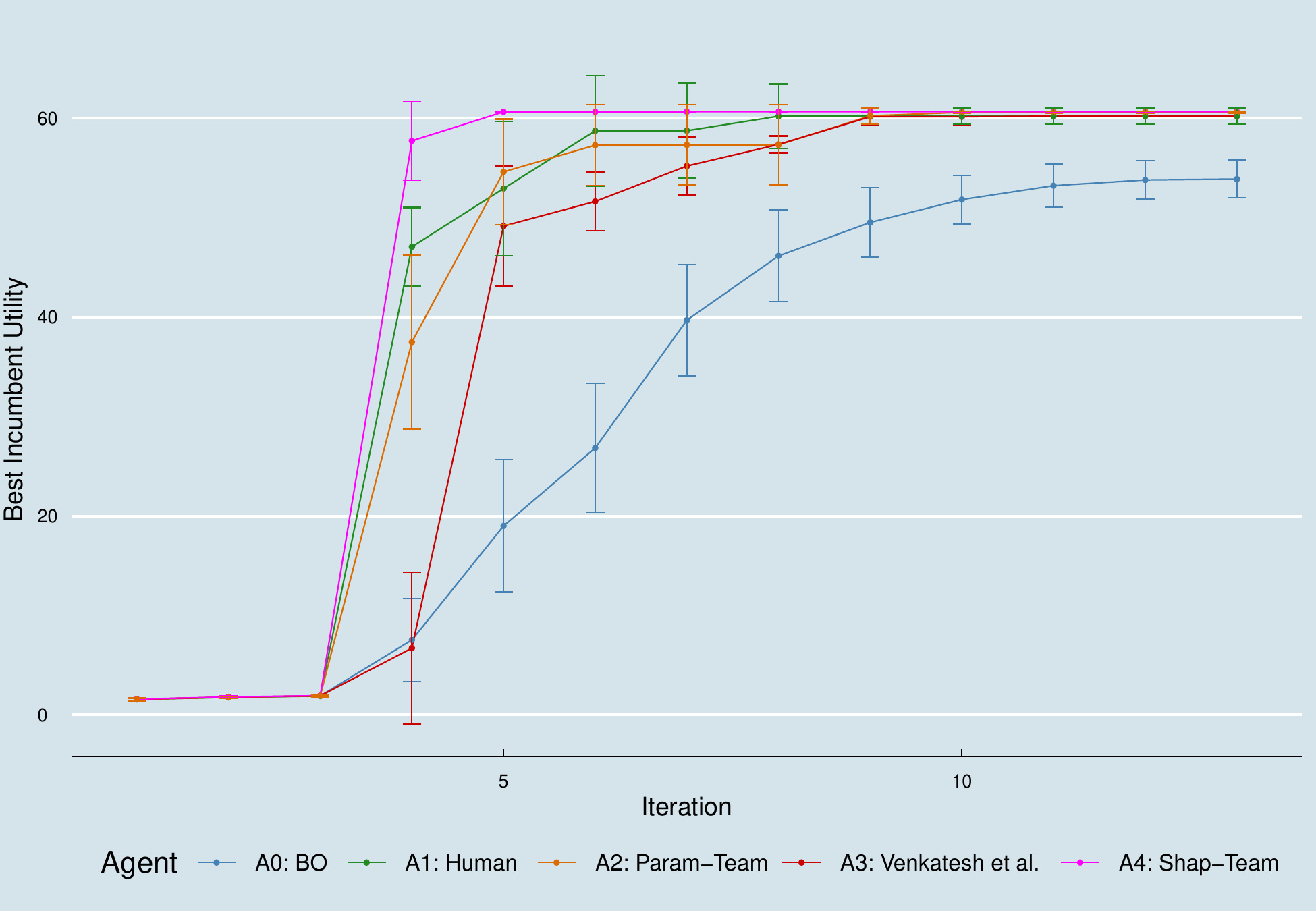}
    \caption{Results of Agents A0-A4 (see Table~\ref{app-tab:my_label}) in human-AI collaborative BO for simulated exosuit personalization (individual 13) with 10 iterations and 3 initial samples each. Error bars indicate $95 \%$ confidence intervals; $k=2$ for A3, $\beta = 2$ for A2 and A4.}
    \label{app-fig:S13}
\end{figure}

\begin{figure}[H]
    \centering
    \includegraphics[width=0.98\linewidth]{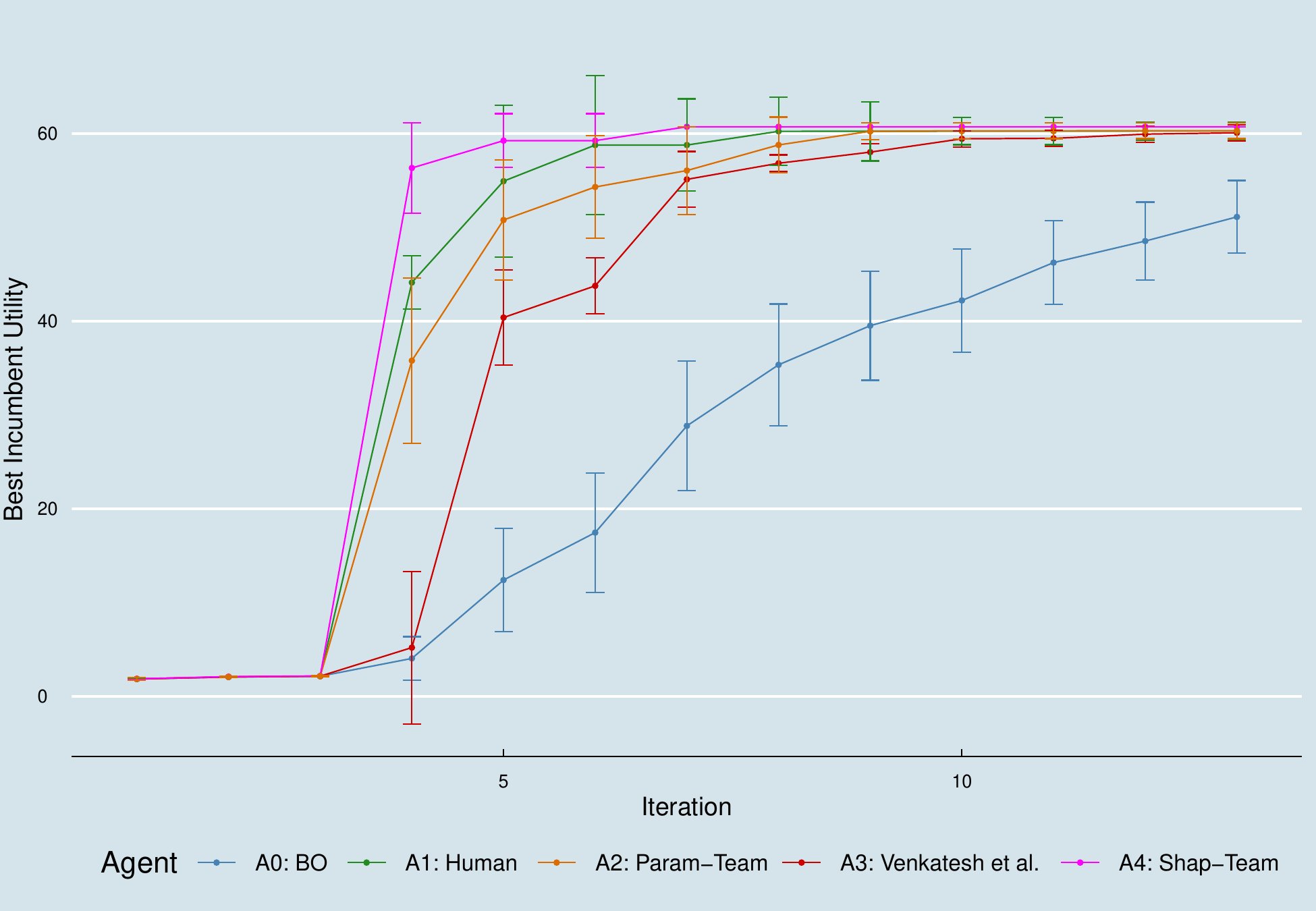}
    \caption{Results of Agents A0-A4 (see Table~\ref{app-tab:my_label}) in human-AI collaborative BO for simulated exosuit personalization (individual 14) with 10 iterations and 3 initial samples each. Error bars indicate $95 \%$ confidence intervals; $k=2$ for A3, $\beta = 2$ for A2 and A4.}
    \label{app-fig:S14}
\end{figure}

\begin{figure}[H]
    \centering
    \includegraphics[width=0.98\linewidth]{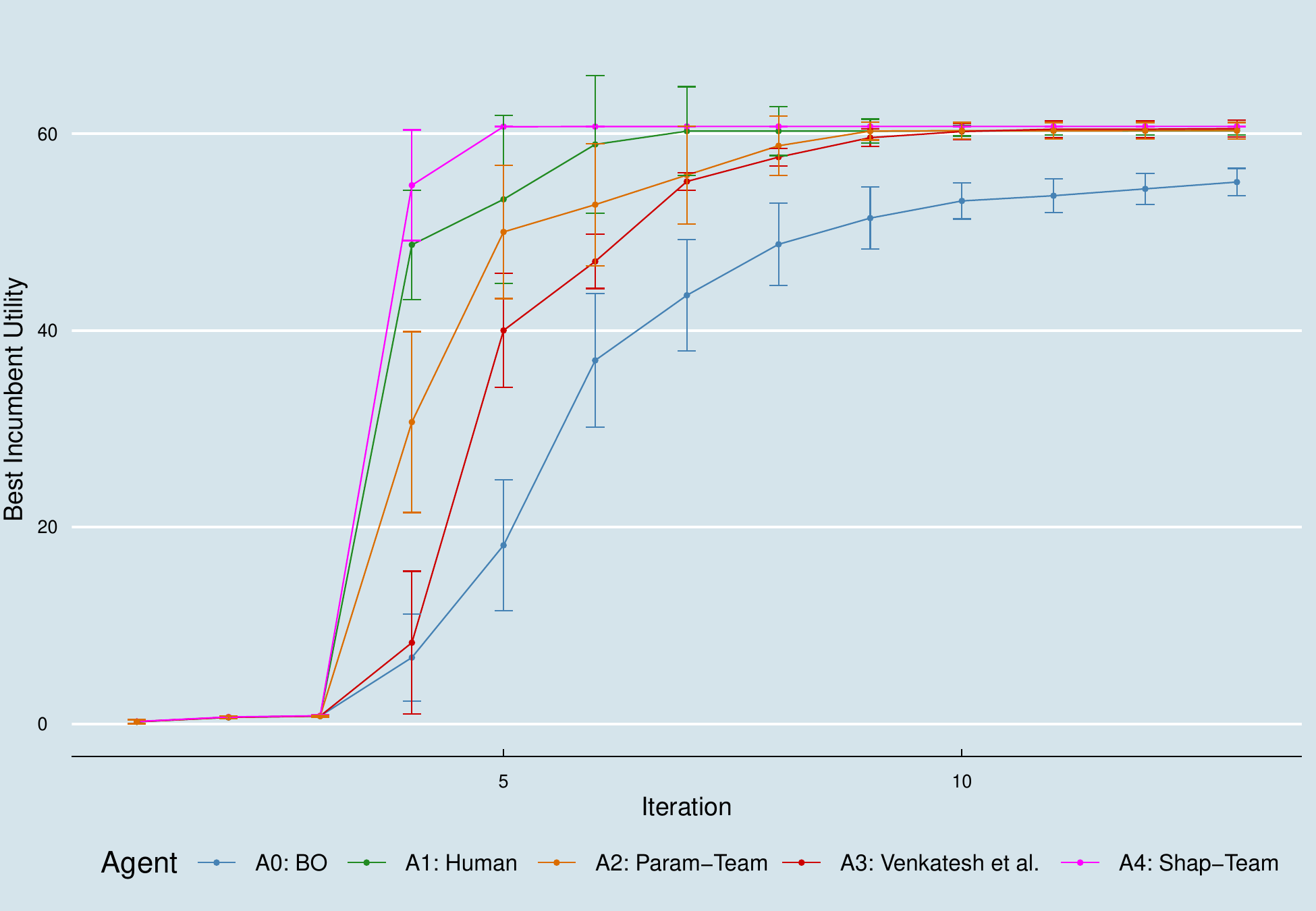}
    \caption{Results of Agents A0-A4 (see Table~\ref{app-tab:my_label}) in human-AI collaborative BO for simulated exosuit personalization (individual 15) with 10 iterations and 3 initial samples each. Error bars indicate $95 \%$ confidence intervals; $k=2$ for A3, $\beta = 2$ for A2 and A4.}
    \label{app-fig:S15}
\end{figure}

For ease of exposition, we only included detailed results for individual 1 in the main paper. Results for remaining 14 individuals are displayed by Figures~\ref{app-fig:S2} through~\ref{app-fig:S15} in what follows. As stated in the main paper, \texttt{ShapleyBO}-assisted human-BO teams outperform competitors on average for all individuals. For 10 out 15 individuals, this finding is significant at confidence level of $95\%$  \\

\textbf{Individual 2:}%

\textbf{Individual 3:}
\newpage

\textbf{Individual 4:}

\textbf{Individual 5:}

\newpage
\textbf{Individual 6:}

\textbf{Individual 7:}

\newpage
\textbf{Individual 8:
}%

\textbf{Individual 9:}

\newpage

\textbf{Individual 10:}

\textbf{Individual 11:}

\newpage
\textbf{Individual 12:}

\textbf{Individual 13:}

\newpage

\textbf{Individual 14:}

\textbf{Individual 15:}

\newpage\null\thispagestyle{empty}\newpage

\bibliography{ijcai24}

\end{document}